\documentclass[journal]{IEEEtran}   

\IEEEoverridecommandlockouts                              

\usepackage{graphicx}      
\usepackage{amsfonts,amssymb,amstext,verbatim,amsmath,color,graphicx,subfigure}
\usepackage{xcolor}
\usepackage{algorithmic}
\usepackage{enumitem}
\usepackage{multirow}
\usepackage{caption} 
\usepackage{cancel}

\usepackage[export]{adjustbox}

\usepackage[linesnumbered,vlined,ruled,boxed,commentsnumbered]{algorithm2e}
\SetKwRepeat{Do}{do}{while}

\def\b#1{\mathchoice{\hbox{\boldmath $\displaystyle #1$}}
        {\hbox{\boldmath $\textstyle #1$}}
        {\hbox{\boldmath $\scriptstyle #1$}}
        {\hbox{\boldmath $\scriptscriptstyle #1$}}}
\newcommand{\nat}{\mathbb{N}}
\newcommand{\rea}{\mathbb{R}}
\newcommand{\preset}[1]{\ensuremath{\,\!^\bullet{#1}}}
\newcommand{\postset}[1]{\ensuremath{{#1}^\bullet}}

\newcommand{\N}{{\ensuremath{\mathcal{N}}}}    
\newcommand{\Pre}{\b{Pre}}                  
\newcommand{\Post}{\b{Post}}                

\newtheorem{theorem}{\textbf{Theorem}}[section]

\newtheorem{example}[theorem]{\textbf{Example}}

\newtheorem{definition}[theorem]{\textbf{Definition}}
\newtheorem{problem}[theorem]{\textbf{Problem}}

\begin{document}

\title{\LARGE \bf On Multi-Robot Path Planning Based on \\Petri Net Models and LTL specifications} 

\author{Sofia Hustiu, Cristian Mahulea, Marius Kloetzer, and Jean-Jacques Lesage
\thanks{S. Hustiu and M. Kloetzer are with the Dept. of Automatic Control and Applied Informatics, Technical University ``Gheorghe Asachi" of Iasi, Romania
\{\{hustiu.sofia,kmarius\}@ac.tuiasi.ro\}.} \thanks{C. Mahulea is with
the Arag\'on Institute of Engineering Research (I3A), University of
Zaragoza, Maria de Luna 1, 50018 Zaragoza, Spain
\{cmahulea@unizar.es\}.} \thanks{J.J. Lesage is with Universit\'e Paris-Saclay, ENS Paris-Saclay, LURPA, 91190, Gif-sur-Yvette, France \{jean-jacques.lesage@ens-paris-saclay.fr\}} \thanks{The work of C. Mahulea has been partially supported by the MINECO ``Salvador de Madariaga" program. This work has been partially supported by  CICYT-FEDER project PID2021-125514NB-I00 at University of Zaragoza.}
}

\maketitle

\begin{abstract} 

This work considers the path planning problem for a team of identical robots evolving in a known environment. The robots should satisfy a global specification given as a Linear Temporal Logic (LTL) formula over a set of regions of interest. The proposed method exploits the advantages of Petri net models for the team of robots and B\"uchi automata modeling the specification. The approach in this paper consists in combining the two models into one, denoted \textit{Composed Petri net} and use it to find a sequence of action movements for the mobile robots, providing collision free trajectories to fulfill the specification. The solution results from a set of Mixed Integer Linear Programming (MILP) problems. The main advantage of the proposed solution is the completeness of the algorithm, meaning that a solution is found when exists, this representing the key difference with our previous work in \cite{kloetzer2020path}. The simulations illustrate comparison results between current and previous approaches, focusing on the computational complexity.
\end{abstract}

\begin{IEEEkeywords}
High-level specifications, formal methods, Petri nets, autonomous robots
\end{IEEEkeywords}


 \section{Introduction}

The significance of developing path planning methods for a team of mobile robots has been intensified in the last years. Many researchers propose different approaches, based on various representations which capture the robots movements in the environment such as transition systems \cite{tumova2016multi,yu2021distributed} or Petri net models \cite{montijano2021probabilistic,ARLaLi19}. In multiple scenarios, the robots are required to fulfill a defined global goal. The most used formalisms to express missions for the team of robots are based on high-level specification, such as: Boolean specifications \cite{mahulea2020multi} and Linear Temporal Logic (LTL) specifications \cite{cohen2021model}. The motion planning must ensure the given mission by computing collision free trajectories for the team members.

Certainly, the association between the environment model and the specification one can be computed in various ways to return \emph{the best} solution for a considered scenario. In previous works, one can notice the complexity drawback of the centralized approach based on transition system representations. Thus, for large teams of robots, the discrete centralized model used to solve the path planning problem generates an exponential increase of the discrete states with respect to the number of robots. This is due to the fact that the centralized model is obtained by the synchronous product of a number of transition systems equal to the number of robots, each transition system capturing the evolution of a robot. Additionally, the resulted model is further extended by doing the synchronous product with the B\"uchi automaton modeling the specification. To overcome this complexity issue, in \cite{tumova2016multi}, the path planning is computed by assuming a different transition system for each robot and the global specification is \emph{distributed} into individual tasks. Each individual task is solved using the transition system of only one robot, hence with a smaller number of states than the centralized model. However, not all LTL formulas can be distributed and the distribution algorithm has a high computational complexity. 

Another approach to overcome the centralized solution using transition system models is based on using a Petri net model of the environment \cite{kloetzer2020path}. The structure of this Petri net model is independent with respect to the number with robots, since the robots are represented by the tokens. Therefore, the models used in the work \cite{kloetzer2020path} are: a Petri net model for the team and the B\"uchi automaton for the LTL formula. Then, a set of $k$ runs in the form of \emph{prefix} and \emph{suffix} in B\"uchi automaton is computed by using the k-shortest path algorithm. The next step is to take a run and try to follow it by obeying the PN model structure. If the run cannot be followed because the observations cannot be generated by the robots or the observations can generate other transitions in B\"uchi, the procedure considers another run and iterates the algorithm. The main problem of this approach is that the algorithm is not complete and it cannot ensure that a solution is obtained (if there exists), but also a set of $k$ runs should be computed.


The main contribution of the current work is to propose a complete algorithm which ensures the attainment of a solution when this is achievable (expressed as collision free robot trajectories). For this reason, the technique aims to use both Petri net model and B\"uchi automaton in a joined model denoted \textit{Composed Petri net} model. The new model exploits the search of an accepted run in B\"uchi while providing feasible independent trajectories in a reduced Quotient Petri net model of the environment. The latter representation decreases the size of the original PN model, considering one place for each unique observation. The complete solution is returned as a result of two MILPs: (1) for joined model denoted \textit{Composed Petri net} model and (2) to project the solution in the original PN model. Furthermore, the complexity of this work depends on the total number of places of the joined model represented by a sum and not a Cartesian product, e.g., automaton product \cite{ding2011automatic}. 

The paper follows the next structure: Section \ref{sec:related} captures several approaches in the field of robot motion planning, revealing the current challenges based on the related work. Section \ref{sec:pbdef} asserts the problem definition, while Section \ref{sec:notations} seizes the mathematical notations used alongside the paper. The complete solution partitioned into its two main steps is explained in Sections \ref{sec_solution}, \ref{sec:project}, joined by examples. Section \ref{sec:comp_simul} is dedicated to the analysis of the results based on several numerical simulations. Also, a comparison with an existing sequential method is illustrated. The last section exposes the conclusions of the current work, while mentioning future improvements. 

\section{Related work}\label{sec:related}

Path planning for mobile robots started by analysing various methods to reach a given destination point from an initial point, especially for single robot scenario \cite{lavalle2006planning}. As the problem is extended towards multi-robot navigation, the focus is directed to concepts such as: (1) type of approach - centralized, decentralized, distributed; (2) the problem formulation based on global or individual mission given for the team of robots. 

High-level specifications can be used to specify the mission for a team of robots (which are usually denoted as {\it agents} to enlarge the aim of the proposed method in other areas). It is often assumed that the robots are identical. One intuitive language is represented by Linear Temporal Logic (LTL), encoding human representation into a logical and temporal formalism, e.g., ``visit region A, then region B and always avoid region C``. Any LTL specification can be modeled as a Rabin automaton \cite{esparza2018one}, or a B\"uchi automaton \cite{kloetzer2020path} for example. In addition, this formalism is efficient in defining a global mission for the entire team \cite{hustiu2020distributed, hustiu2021ltl}, or local mission for each agent \cite{guo2015multi}. The LTL formalism is effective also to plan optimal motion trajectories \cite{wolff2014optimization}. In \cite{hustiu2020distributed}, the global LTL mission is decomposed into individual independent tasks, for a flexible centralized approach of task assignment with respect to the robots. This work was extended afterwards in the 3D space, for a team of Unmanned Aerial Vehicles (UAVs) \cite{hustiu2021ltl}. On the other hand, the paper \cite{guo2015multi} is directed to solve the path planning problem by assigning individual LTL tasks for each robot. The robots cooperate while satisfying hard (for collision free trajectories) and soft LTL constraints. The notion of LTL soft and hard constraints is also included in others works, e.g., \cite{dimitrova2018maximum}.

There are various mechanisms to combine the LTL specification model with the representation of the dynamic robotic system. One approach is based on transition system models, based on a partitioned environment (defining the nodes) and their adjacency relation (defining the edges) \cite{mahulea2020path}. In \cite{yu2021distributed}, the authors use the transition system model for a group of mobile robots, with individual scope given for each robot. A different approach is illustrated in \cite{tabuada2005motion}, where the nodes illustrate the kinematic of individual agents and the edges capture the inter-agent constraints. All of these approaches encounter the same disadvantage as a result of using the transition system to model the dynamic robotic system, this being expressed as the exponential increase of complexity. This is the result of automaton products \cite{ding2011automatic}. To overcome this downside, the paper \cite{tumova2016multi} reduces the complexity of multi-robot path planning by considering the following: decomposing the problem into multiple finite horizon planning issues, and solving them in iterative manner, using an event-based synchronization between agents. 

Another solution to overcome the state explosion problem is by using another representation for the motion of the robots in the workspace, such as Petri net (PN) model. Considering a given environment, PN model has a fixed topology subject to the number of robots, and is easily adaptable related to their initial position. In \cite{mahulea2020multi} the collision free trajectories of robots are obtained as a result of solving two Mixed Linear Integer Programming (MILP) problems, while the environment is modeled as a PN and a Boolean specification is given. The approach used in \cite{costelha2012robot} consists in the use of three layered PN model which represent the environment, the changes in the work-space, and the task plan based on the compositions different types of events and actions towards goal. 

The benefits of the Petri net models are illustrated also in works which consider unknown or partially known environment, e.g., \cite{montijano2021probabilistic} - the robots need to avoid regions which have a high probability of collision, while satisfying a Boolean specification, without knowing their precise location.

So far we have presented the benefits of using both models to achieve collision free trajectories while a given specification should be fulfilled: PN models for the dynamic robotic system and LTL specification for the team's mission. A ``high-level" view of mobile planning can be catalogued as follows: in \cite{lacerda2011ltl} a PN model is assigned to each robot, ensuring a global LTL mission; in \cite{dai2016cooperative} - each robot is modeled as an automaton, being supervised periodically; in \cite{lacerda2019petri} - capturing the notion of supervisory control, where a framework for multi-robot system is enforced.

On the other hand, from a ``low-level" point of view it is observed that is not easy to design a controller for a team of agents satisfying a given mission. Several works exploit these models separately, omitting some advantages which could appear by using both models in the same time. One such example is proposed in paper \cite{lacerda2019petri} considering a parallel execution of transitions in the PN model (of the workspace) and the B\"uchi automaton (of the LTL formula), based on the construction of a new PN supervisor model. In other words, one transition in the environment model is fired only if a transition in the B\"uchi automaton is satisfied. The authors mention that is difficult to design a PN supervisory model for complex tasks, the construction being prone to errors especially without having a formal method of verification. We proposed a different approach in a previous work \cite{kloetzer2020path}: the B\"uchi automaton and the PN model are used sequentially compared with the parallel approach from \cite{lacerda2019petri}. Herein, a single run in the B\"uchi automaton is pursued, followed by a sequence of the robots in the PN model with respect to the selected run. The procedure is iterated until the robots meet the LTL specification. While the results demonstrate the benefit of this approach, the proposed algorithm is not complete. 

With this work we propose a method based on a joined \textit{Composed Petri net} model. In this sense, a Quotient PN model is computed with respect to the original PN model of the environment, where each place models an unique observation. The reduced model of the dynamic robotic system is afterwards combined with the B\"uchi automaton. The movement sequence for the team of robots is returned by the solution of two MILPs (based on the reduced model and a projection in the original PN model), their trajectories satisfying a global LTL specification. To the best of our knowledge, this composition of Petri net and B\"uchi automaton was never done before to solve such path planning problems. The proposed algorithm is complete, while the complexity increase is reduced due to a smaller number of places in the joined Petri net model. Compared with other approaches based on automaton product which suffer from an exponential increase in the number of states, the design of \textit{Composed Petri net} model handles a number of places equal to a sum between places in the Quotient PN, number of states in B\"uchi automaton and twice the number of individual observations.

\section{Problem statement}\label{sec:pbdef}
Let us consider the next problem definition: a set of identical robots denoted $R=\{r_1, r_2, \ldots, r_{|R|}\}$ evolves in an known and static 2D environment. The robots are assumed omnidirectional and of negligible size, i.e., each is a fully actuated point. The environment captures several convex polygonal shapes, denoted as regions of interest (ROI) and labeled with elements from the set $\mathcal{Y} = \{y_1,y_2,\ldots, y_{|\mathcal{Y}|} \}$. These regions can be both disjointed and overlapped. Furthermore, based on these polygonal shapes, we assume that the environment is partitioned into a set of regions also called cells, for example by using a cell decomposition method \cite{mahulea2020path,Choset05}. Each cell entirely belongs to the same region(s) of interest, or it is included in the area not covered by any region of interest (free space). The advantage of this representation lies in obtaining an abstract model of the environment, e.g., Petri net model, which captures the entire space data, being invariant with respect to the number of robots.

The set of cells is denoted by $P=\{p_1, p_2, \ldots, p_{|P|}\}$ and the set of labels assigned to a cell is given by a function $h :P\rightarrow \mathcal{Y}\cup\{\emptyset\}$. If the cell $p_i$ is included in the intersection of ROIs $y_j$ and $y_k$, then $h(p_i) = \{y_j, y_k\}$, while if $p_l$ lies in the free space, then $h(p_l)=\emptyset$. Fig. \ref{fig:fig1}(a) can be used as example for an intuitive illustration.

\begin{problem}\label{def:problem}
Assume that the movement of a robotic team is abstracted to a Robot Motion Petri Net (RMPN) model given in Def. \ref{def:rmpn}. Given a global LTL specification over the set $\mathcal{Y}$, automatically compute trajectories for the robots in order to fulfill the specification.

\begin{example}\label{ex:buchi}
Let us consider the following mission task for the robots in Fig. \ref{fig:fig1}(a): 
\begin{equation} \label{eq:1}
\varphi= \diamondsuit \left( y_1 \wedge y_2 \wedge y_3 \right) \wedge \neg \left( y_1 \vee y_2 \right) \mathcal{U} \left( y_1 \wedge y_2 \right)
\end{equation}

This task means that regions $y_1$, $y_2$ and $y_3$ should be \emph{eventually} visited at the same time, and $y_1$ and $y_2$ should be reached simultaneously. \hfill $\blacksquare$
\end{example}
\end{problem}

The main idea proposed in \cite{kloetzer2020path} computes an accepted run in the B\"uchi automaton $B$, and then searches trajectories in the RMPN in order to sequentially generate observations that follow the imposed run of $B$. However, when the trajectories are computed in the RMPN, some other observations can be generated that could produce a transition in $B$ that deviates from the imposed run. In this case, the procedure is iterated by imposing another accepted run in B\"uchi and then try to follow it with RMPN's outputs. For this approach, several Mixed Integer Linear Programming (MILP) optimization problems were solved. The algorithmic strategy in \cite{kloetzer2020path} is not complete, and it does not account for collision free trajectories, aspects that are solved by the strategy herein, in this paper.

In the following sections we proposed a different solution than the one in \cite{kloetzer2020path} which can be divided in two steps. First, we derive a PN model in which we embed an abstraction (Quotient) of RMPN with the B\"uchi automaton $B$, denoted \textit{Composed Petri net} model. In the second step, a run is found in the new joined Petri net model, with the property of being feasible, while generating a sequence of outputs accepted by $B$. The obtained run is projected on the original RMPN and the robot trajectories are computed. 

For each step in the next sections, the paper is accompanied by mathematical notations and examples, for an easier understating of the proposed algorithm. 

\section{Notations and Preliminaries}\label{sec:notations}


\begin{definition}\cite{mahulea2020path}\label{def:rmpn} A Robot Motion Petri Net system (RMPN) is a tuple $\mathcal{Q} = \langle \N, \b{m}_0, \mathcal{Y}, h \rangle$, where:
\begin{itemize}
\item $\N=\langle P, T, \Post, \Pre \rangle$ is a Petri net with:
\begin{itemize}
    \item The set of places $P$ (one place for each  cell);
    \item Set of transitions $T$, each transition corresponding to a robot movement between adjacent cells;
    \item $\Post \in \{0,1\}^{|P|\times |T|}$ is the post-incidence matrix, defining the arcs from transitions to places, with $\Post[p,t]=1$ if $t\in T$ is connected with place $p\in P$, otherwise $\Post[p,t]=0$;
    \item $\Pre \in \{0,1\}^{|P|\times |T|}$ is the pre-incidence matrix defining the arcs from places to transitions, with $\Pre[p,t]=1$ if place $p\in P$ is connected with transition $t\in T$, otherwise $\Pre[p,t]=0$;
\end{itemize}    
\item $\b{m}_0$ is the initial marking, where $\b{m}_0[p]$ gives the number of robots initially deployed in cell $p\in P$;
\item $\mathcal{Y}\cup \{\emptyset\}$ is the set containing the output symbols;
\item $h:P\rightarrow \mathcal{Y}\cup\{\emptyset\}$ is the observation map, defined above. Thus, if $p_i$ has at least one token (i.e., at least one robot is currently in cell $p_i$), then region(s) of interest $h(p_i)$ is (are) visited.
\end{itemize}
\end{definition}

\begin{figure*}
    \centering
    \begin{tabular}{cc}
    \includegraphics[width=.5\textwidth]{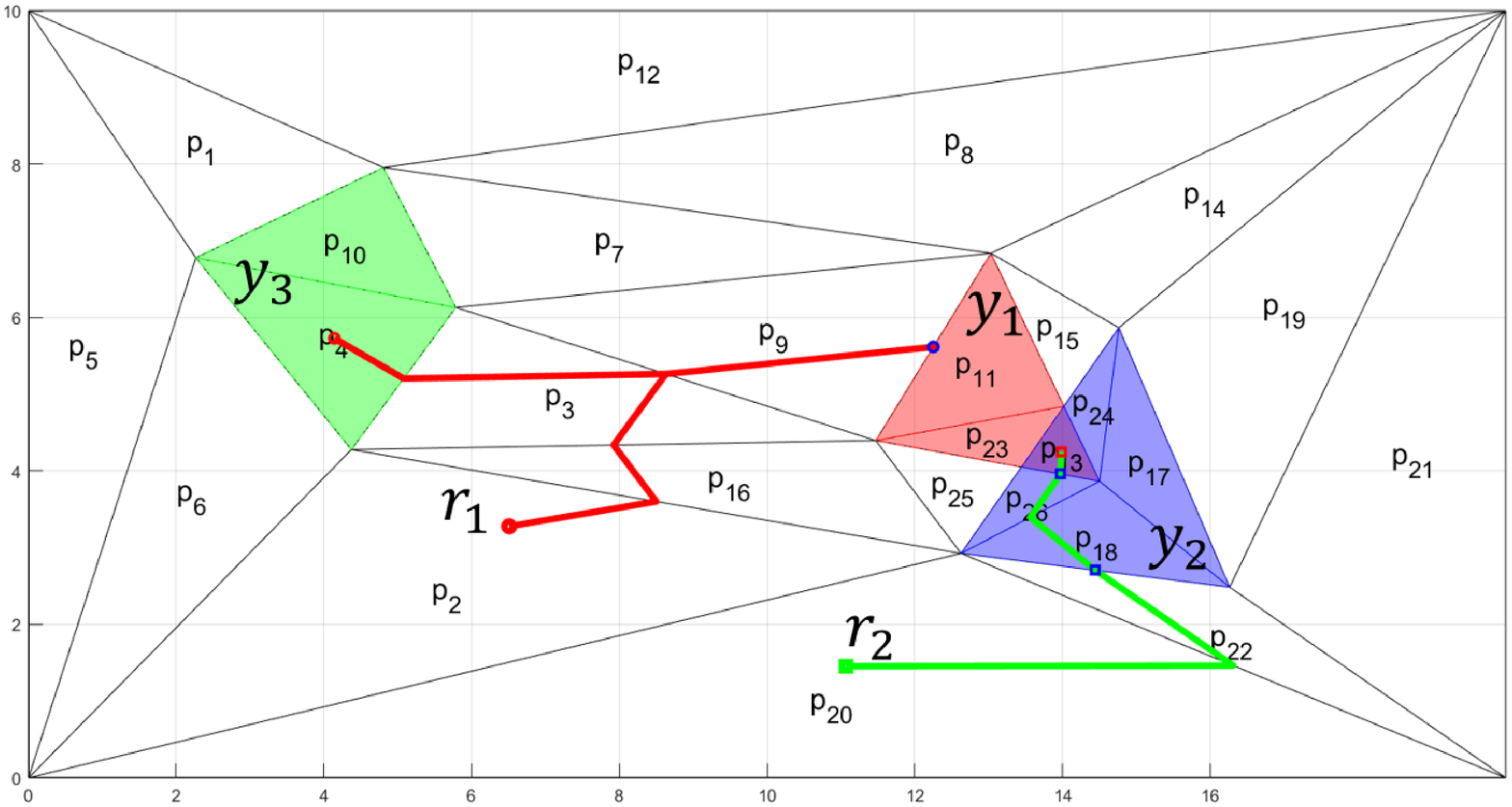} & 
 \includegraphics[width=.45\textwidth]{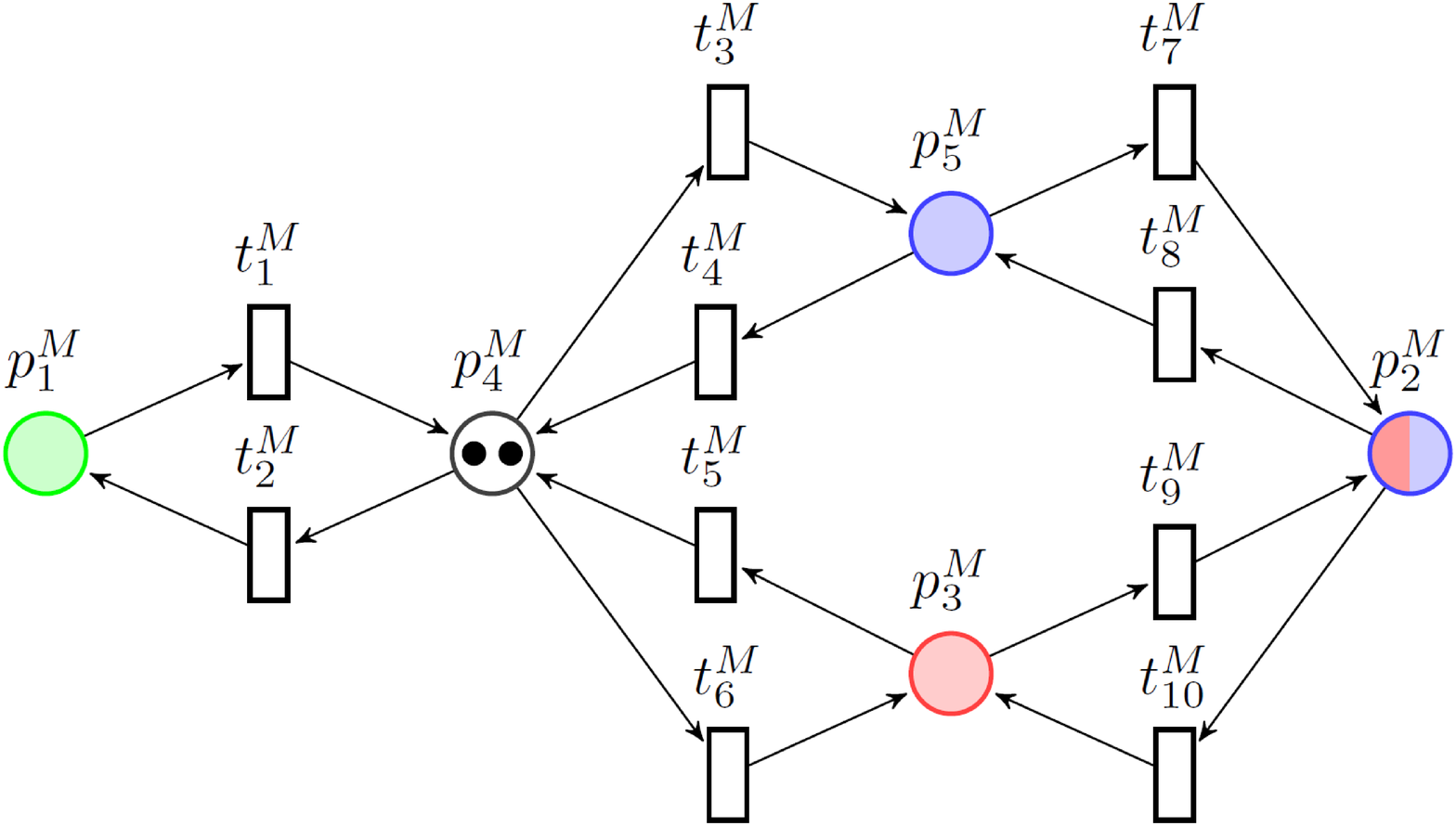}
\\ (a) The environment used in Example \ref{ex:buchi}. & (b) Quotient Petri net of the RMPN. \end{tabular}
    \caption{Example of an environment and the corresponding RMPN's quotient.}
    \label{fig:fig1}
\end{figure*}

\begin{example}\label{ex:env}
Let us consider the environment in Fig. \ref{fig:fig1}(a), containing three ROIs ($\mathcal{Y} = \{y_1,y_2,y_3\}$) and two robots, initially located in $p_2$ and $p_{20}$. Therefore, $h(p_{11}) = h(p_{23}) =y_1$, $h(p_{17}) = h(p_{18}) = h(p_{24})= h(p_{26}) =y_2$, $h(p_{4}) = h(p_{10}) =y_3$, $h(p_{13})=\{y_1, y_2\}$, and $h(p_i) = \emptyset$ otherwise.

The RMPN modeling the movement capabilities of this team of two robots consist in a set of 26 places and 74 transitions. The set of outputs $\mathcal{Y}$ and the observation map $h$ are given before. The initial marking $\b{m}_0$ is a vector of dimension 26 having all elements equal to zero except $\b{m}_0[p_2]=\b{m}_0[p_{20}]=1$.\hfill $\blacksquare$
\end{example}

Each token in the RMPN models the current location (cell) of each robot, hence the total number of tokens is equal to $|R|$. Notice that by using this definition, the structure of the model (number of places, transitions and arcs) is not changing if robots identical with others are added (removed) to (from) the team. Only the marking (state) of the RMPN is changed.


For a generic transition $t_j \in T$, $\preset{t}$ denotes its input place, while $\postset{t}$ denotes its output place\footnote{By definition, RMPN systems considered in this paper belongs to the class of state-machine, i.e., each transition has one input and one output place.}. An enabled transition $t_j$ can fire, and the RMPN reaches a new marking $\b{\tilde{m}} = \b{m} + \b{C}[\cdot, t_j]$, where $\b{C} = \Post - \Pre$ is the token flow matrix and $\b{C}[\cdot, t_j]$ is its column corresponding to $t_j$. Based on the construction of the RMPN, the firing of a transition $t_j$ corresponds to the movement of a robot from cell $\preset{t_j}$ to cell $\postset{t_j}$. For the moving robot, transition $t_j$ means to apply a control law that drives the robot from cell $\preset{t_j}$ to $\postset{t_j}$, and there exist approaches for designing such continuous laws in specific scenarios \cite{HabColSchup06,Belta-TAC06}.

We will be interested in finding sequences of transitions to be fired such that the team fulfills a given specification. If a RMPN marking $\b{\tilde{m}}$ can  be reached from $\b{m}$ through a finite sequence of transitions $\sigma$, we denote with $\b{\sigma} \in \nat^{|T|}_{\geq 0}$ the firing count vector, i.e., its $j^{th}$ element is the cumulative amount of firings of $t_j$. In this case, the state (or fundamental) equation is satisfied:
\begin{equation}\label{eq:steq}
    \b{\tilde{m}} = \b{m} + \b{C} \cdot \b{\sigma}
\end{equation}

A firing vector $\b{\sigma}$ can be found, having a minimum number of transitions that drives the live\footnote{A Petri net is live if independently by the actual reachable marking, all transitions can fire in the future.} RMPN to a desired marking $\b{\tilde{m}}$, i.e., by solving the optimization problem $\min \b{1}^T\cdot \b{\sigma}$. The details of transforming the firing vector into a sequence of robot movements are captured in \cite{mahulea2020path}.

\textbf{LTL formulae and B\"{u}chi automata}
The global motion task that the robot team should fulfill is specified as an $LTL_{-X}$ formula  \cite{Clarke99,baier2008principles}. The subclass of LTL denoted as $LTL_{-X}$ contains formulae recursively defined over a set of atomic propositions $\mathcal{Y}$, by using the standard Boolean operators ($\neg$ - negation, $\wedge$ - conjunction, $\vee$ - disjunction, $\Rightarrow$ - implication, and $\Leftrightarrow$ - equivalence) and some temporal operators (${\cal U}$  - until, $\diamondsuit$ - eventually, and $\square$ - always). For simplicity of notation we further write LTL instead of $LTL_{-X}$, the difference being the absence of the next operator $\bigcirc$ in $LTL_{-X}$ \cite{Belta-RAM07,FainekosGKP09,baier2008principles}.


As proved in \cite{Wolper83}, any LTL formula can be transformed into a non-deterministic B\"uchi automaton that accepts all and only the input strings satisfying the formula. Automatic translation from an LTL formula to B\"uchi automaton can be done by using available tools, for example \cite{Holzmann04,Gastin01,Spot}. The property of LTL being close under stuttering \cite{etessami1999stutter} is exploited in the paper. This means that any finite repetition of the same input does not modify the truth value of the input string. The stuttering is useful in Section \ref{sec:project}, when the obtained solution in the reduced Quotient Petri net model is projected into the original RMPN.

\begin{definition}\label{def:Buchi}
The B\"{u}chi automaton corresponding to an LTL formula over the set $\mathcal{Y}$
has the structure $B=\left(S,S_0,\Sigma_B,\rightarrow_B,F\right)$, where:
\begin{itemize}
\item $S$ is a finite set of states;
\item $S_0\subseteq S$ is the set of initial states;
\item $\Sigma_B$ is the finite set of inputs;
\item $\rightarrow_B\subseteq S\times\Sigma_B\times S$ is the transition
    relation;
\item $F\subseteq S$ is the set of final states.\hfill $\blacksquare$
\end{itemize}
\end{definition}

$B$ can be non-deterministic, by means of enabling multiple outgoing transitions from the same input state, e.g., $(s,\tau,s')\in\rightarrow_B$ and $(s,\tau,s'')\in\rightarrow_B$, with $s'\neq s''$. We denote by $\pi(s_i,s_j)$ the set of all inputs that enable transition from $s_i$ to $s_j$, expressed as a Boolean formula over the set $\mathcal{Y}$, reduced to a Disjunctive Normal Form (DNF). The input can be also expressed as combination of active observations over the set $2^\mathcal{Y}$, where $\emptyset$ represents the free space. To reduce the complexity in the proposed algorithm, the selected formulation is based on Boolean formulae, where a single conjunctive element is denoted with $\alpha_{k}$.

An {\it infinite accepted run} in $B$ drives the automaton towards a final state from an initial state. This run can be stored on finite memory with regards to two finite length strings (i) {\it prefix} - towards a final state in set $F$; (ii) {\it suffix} - towards the same final state reached by the prefix. The run can be written as {\it prefix, suffix, \ldots}. More about infinite runs in $B$, joined by computing the sequence of elements from $\Sigma_B$ can be found in \cite{Wolper83}.



\begin{figure*}
    \centering
    \begin{tabular}{cc}
    \includegraphics[width=.3\textwidth]{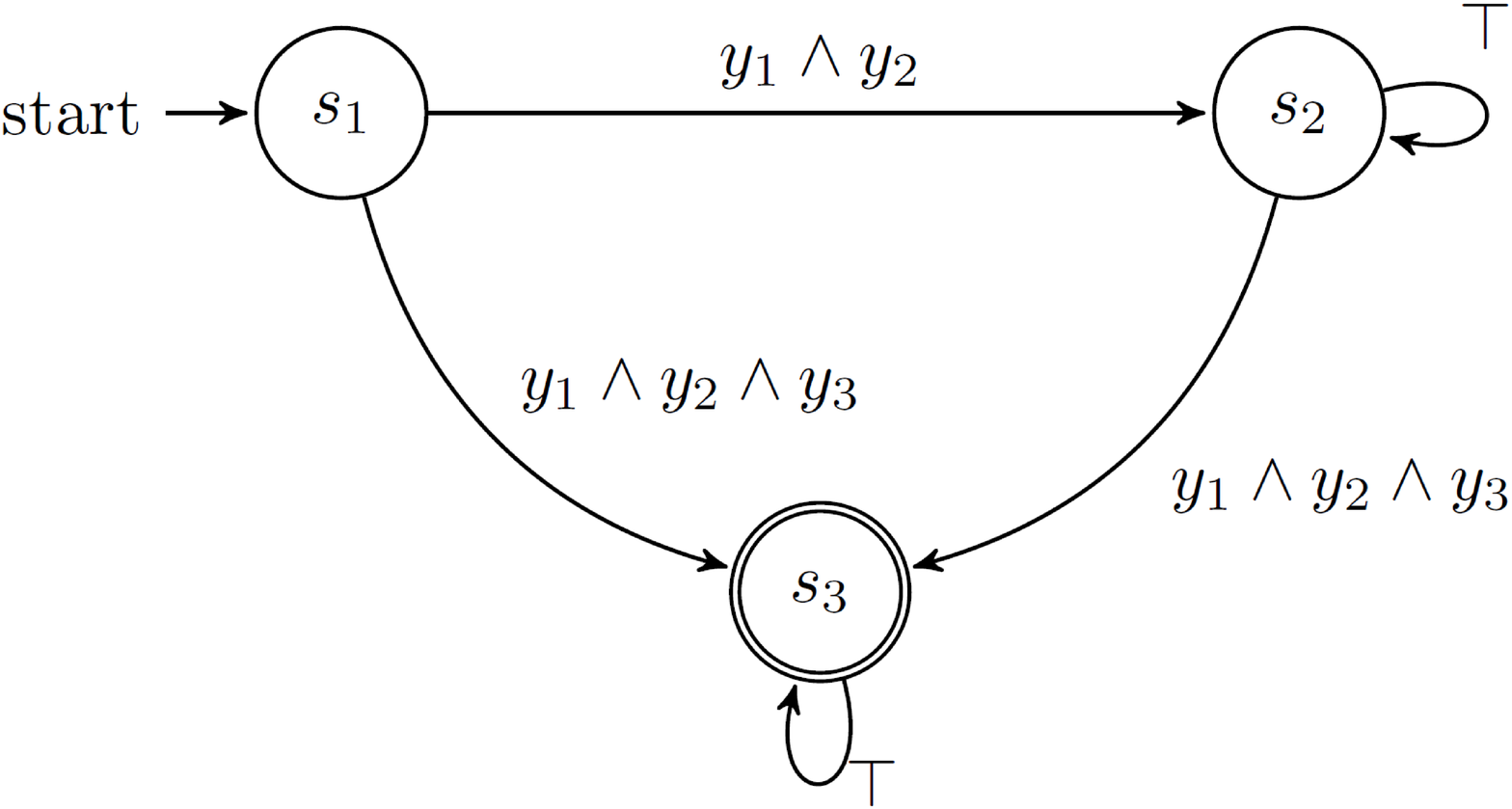} & 
 \includegraphics[width=.3\textwidth]{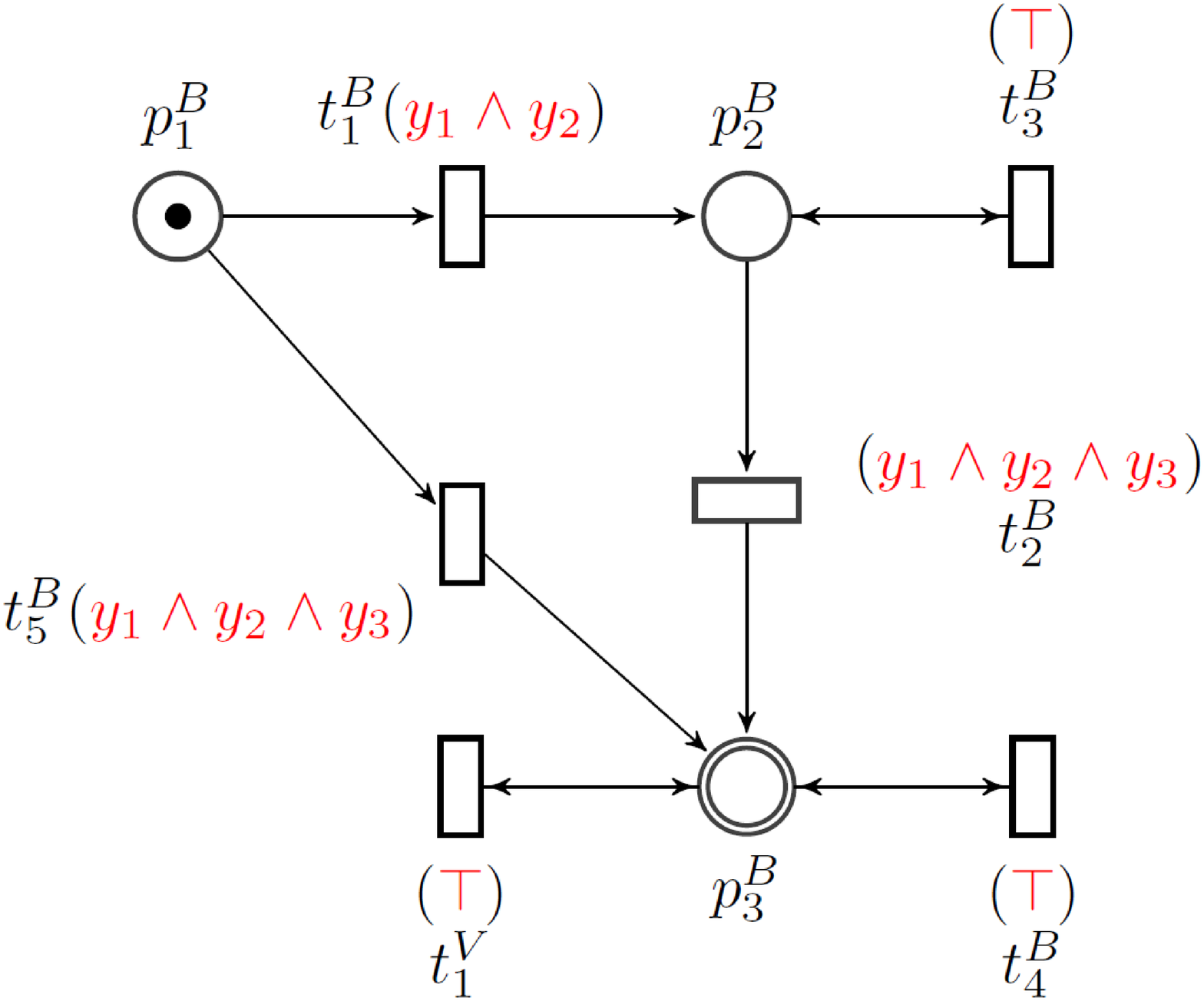}
\\ (a) B\"uchi automaton for the LTL formula in Ex. \ref{ex:buchi}. & (b) B\"uchi Petri net corresponding to the B\"uchi automaton in Fig. \ref{fig:fig2}(a) \end{tabular}
    \caption{Example of B\"uchi automaton and B\"uchi Petri net}
    \label{fig:fig2}
\end{figure*}

\begin{example}\label{ex:descBuchi}
Let us recall the Example \ref{ex:buchi}. The B\"uchi automaton corresponding to this LTL task is given in Fig. \ref{fig:fig2}(a), where symbol $\top$ (True) means any observation from $2^\mathcal{Y}$. On the other hand, Fig. \ref{fig:fig2}(b) illustrates the B\"uchi Petri net model, having the Boolean formula of transitions represented with red color. This model will be described in Section \ref{sec_solution}. An accepted run satisfying the formula could be $s_1,\, s_3,\, s_3, \ldots$ with prefix $s_1$ and suffix $s_3$. Notice that the unique final state $s_3$ is visited infinitely often. However, as it may be observed in the environment of Fig. \ref{fig:fig1}(a), this run cannot be generated by the two robots. Indeed, the input necessary for the first run's transition of is $\pi(s_1,s_3)=y_1 \wedge y_2 \wedge y_3$, which cannot be generated when starting from the initial location of the robots ($p_2$ and $p_{20}$). This is because the two robots should evolve only through cells belonging to free space until one enters $p_{13}$ and the other simultaneously enters $p_4$ or $p_{10}$. Clearly, entering in $p_{13}$ directly is not possible since it would require to activate first $y_1$ or $y_2$ and would violate $\varphi$. 

A possible run that can be generated by the robots is: $s_1,\, s_2,\, s_3,\, s_3, \ldots$ with prefix $s_1,\, s_2$ and suffix $s_3$. For this, the robots should first enter in $y_1$ and $y_2$ synchronously (generating $\pi(s_1,s_2)=y_1 \wedge y_2$), then one robot should go to $p_{13}$ (the intersection of $y_1$ and $y_2$). After this, the other robot will move to the free-space, and the output generated by the team remains in $y_1, y_2$, no matter in what cell from free-space is located the second robot. Finally, the robot from the free-space should enter to $y_3$, the team generates output $\pi(s_2, s_3)=y_1 \wedge y_2 \wedge y_3$, and the transition to $s_3$ in B\"uchi automaton is enabled. The robots can stop in these final regions since the self-loop in $s_3$ includes all possible observations.

Let us remark that the solution is not unique, since the self-loop transition in $s_2$ can be taken with any possible input. In particular, the robot that is going to $p_{13}$ can wait in the previous cell and enter in $p_{13}$ synchronously with the other robot in $y_3$. The above intuitive explanations lead to the idea of automatically obtain a team movement strategy which satisfies the LTL formula, thus formulating the problem we solve.\hfill $\blacksquare$
\end{example}

Various notations are used in describing the proposed algorithm for the complete solution, divided into several steps. These symbols are significant when expressing the tuple $\mathcal{Q}$ for RMPN, next to its components. The Table \ref{table:notations} captures a summarized description for an established overview. In this sense we propose a general notation for all the components denoted as ${< \cdot >}$, accompanied by different symbols for different topics. The detailed explanations are included in each section alongside the introduced equations and pseudo-codes. 

\begin{table}[h!]
\centering
\begin{tabular}{||c c||} 
 \hline
 \multirow{1}{6em}{Notation} & \multirow{1}{20em}{Description} \\ [0.5ex] 
 \hline\hline
\multirow{2}{6em}{${< \cdot^M >}$} & \multirow{2}{20em}{Denotes the variables used for Quotient RMPN (Sub-step 1.1)}  \\ [2.5ex] 
\multirow{2}{6em}{${< \cdot^B >}$} & \multirow{2}{20em}{Denotes the variables used for B\"uchi RMPN (Sub-step 1.2)} \\ [2.5ex] 
\multirow{2}{6em}{$< \cdot^C >$} & \multirow{2}{20em}{Denotes the variables used for \textit{Composed Petri net} (Sub-step 1.3)} \\ [3.5ex]
 \hline 
\end{tabular}
\caption{Notations for various PNs to be used}
\label{table:notations}
\end{table}

\section{Solution step 1: Full Petri net model}\label{sec_solution}

\begin{figure}
    \centering
    \includegraphics[width=.6
    \columnwidth]{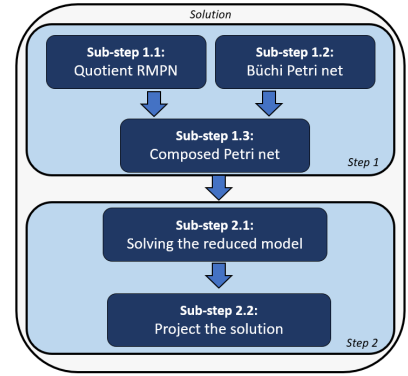}
    \caption{Diagram for the global algorithm}
    \label{fig:completealg}
\end{figure}

Fig. \ref{fig:completealg} captures the steps and sub-steps of the method we propose to achieve the final solution to the stated Problem \ref{def:problem}. The first phase is responsible to output the full Petri net model. In this sense, two Petri net models are assigned to (i) the environment (Sub-step 1.1) - considering a reduced abstraction of the entire space denoted Quotient Petri net and to (ii) the LTL specification (Sub-step 1.2) - based on the B\"uchi automaton. The composition of these models is performed (Sub-step 1.3) in a reduced representation which accounts the active observations while ensuring the given specification. The second phase focuses on the final solution. For this, two actions are required: Sub-step 2.1 returns the solution based on the \textit{Composed Petri net} model, while Sub-step 2.2 projects this solution into a sequence of robot trajectories. 

In this section we employ first a \emph{Quotient} model of the RMPN presented in \cite{IPViMaKl17} that is obtained by aggregating adjacent cells with identical observations. 

\textbf{Sub-Step 1.1. Quotient of the RMPN in Def. \ref{def:rmpn}}. Given a RMPN system $\mathcal{Q}$ as in Def. \ref{def:rmpn}, the idea of obtaining the quotient $\mathcal{Q}^M$ is to iteratively combine any places $p_i$ and $p_j$ from $P$ that satisfy $p_j\in\postset{\left( \postset{p_i}\right)}$ and $h(p_i)=h(p_j)$. This reduction technique is synthesized in Alg. \ref{alg:reduce_pn}, and the reduced PN model $\mathcal{Q}^M$ has the property that its output is updated when a transition is fired. Therefore, after the firing of one transition in $\mathcal{Q}^M$, one transition in B\"uchi should be fired too. Moreover, since a transition of $\mathcal{Q}^M$ corresponds to a set of trajectories in $\mathcal{Q}$, any sequence of transition firings in $\mathcal{Q}^M$ can be translated to a run in $\mathcal{Q}$.

The main loop in Alg. \ref{alg:reduce_pn} (lines \ref{alg_quotient_loop} - \ref{alg_quotient_end_loop}) is iterated until there exist no adjacent places with the same observation. In each iteration, for any pair of adjacent places $\langle p_i^M, p_j^M \rangle$ with the same observation, the two transitions $t_k^M$ and $t_l^M$ modeling the movements of a robot from $p_i^M$ to $p_j^M$ and from $p_j^M$ to $p_i^M$ are removed (lines \ref{alg_quotient_com_ts} - \ref{alg_quotient_rem_ts}). Then, $p_i^M$ and $p_j^M$ are fused in $p_i^M$ by updating the marking vector, the incidence matrices (lines \ref{alg_quotient_fuse} - \ref{alg_quotient_end_fuse}) and the projection matrix $\b{Pr}$ (lines \ref{alg_quotient_up_s} - \ref{alg_quotient_up_fin_s}).

\begin{example}
Let us consider the RMPN for the environment captured in Fig. \ref{fig:fig1}(a). By aggregating the states with the same observation (applying Alg.~\ref{alg:reduce_pn}), the Petri net system $\mathcal{Q}^M$ in Fig. \ref{fig:fig1}(b) is obtained. This new PN model is also a RMPN according to Def.~\ref{def:rmpn}, where each place corresponds to a set of regions in the original RMPN system. Notice that Alg.~\ref{alg:reduce_pn} also returns the projection matrix $\b{Pr}$, in this case of size $5 \times 26$ and having all elements equal to zero except the following:

\noindent $\bullet \b{Pr}[p_1^M,p_4]=\b{Pr}[p_1^M,p_{10}] = 1$ saying that place $p_1^M$ combines $p_5$ and $p_{10}$;

\noindent $\bullet \b{Pr}[p_3^M, p_{11}]=\b{Pr}[p_3^M, p_{23}]=1$;
    
\noindent $\bullet \b{Pr}[p_5^M,p_{17}]=\b{Pr}[p_5^M, p_{18}]=\b{Pr}[p_5^M, p_{24}]= \b{Pr}[p_5^M, p_{26}]=1$;

\noindent $\bullet \b{Pr} [p_2^M,p_{13}]=1$;

\noindent $\bullet \b{Pr}[p_4^M, p_i] = 1$ for all $p_i \in P$ with $h(p_i)=\emptyset$.\hfill $\blacksquare$
\end{example}

\begin{algorithm}[h]
\caption{Quotient of RMPN system}
\label{alg:reduce_pn}
\KwIn{$\mathcal{Q} = \langle \langle P, T, \b{Pre}, \b{Post}\rangle, \b{m}_0, {\cal Y}, h \rangle$}
\KwOut{$\mathcal{Q}^M = \langle \langle P^M, T^M, \b{Pre}^M, \b{Post}^M \rangle, \b{m}^M_0, {\cal Y}, h \rangle$, $\b{Pr}$}

 $P^M = P;\ T^M=T$\; 
 $\b{Pre}^M=\b{Pre}$; $\b{Post}^M=\b{Post}$;
 $\b{m}^M_0 = \b{m}_0$\;
 $\b{Pr} = \b{I}^{|P| \times |P|}$ \tcc{$\b{Pr}$ is the projection matrix}

\While{$\exists p_i^M, p_j^M \in P^M$ such that $p_j^M \in \postset{\left( \postset{p_i^M}\right)}$ and $h(p_i^M)=h(p_j^M)$ \label{alg_quotient_loop}}{
    Let $t_k^M = \postset{p_i^M} \cap \preset{p_j^M}$ and $t_l^M = \preset{p_i^M} \cap \postset{p_j^M}$\; \label{alg_quotient_com_ts}
    Remove columns corresponding to $t_k^M$ and $t_l^M$ from $\b{Pre}^M$ and $\b{Post}^M$\;
    $T^M = T^M \setminus \{t_k^M, t_l^M\}$\; \label{alg_quotient_rem_ts}
    $\b{m}^M_0[p_i^M] = \b{m}^M_0[p_i^M] + \b{m}^M_0[p_j^M]$\;\label{alg_quotient_fuse}
    Remove row $p_j^M$ from $\b{Pre}^M$ and $\b{Post}^M$\;
    Remove element $p_j^M$ from $\b{m}^M_0$\;
    $P^M = P^M \setminus \{p_j^M\}$\; \label{alg_quotient_end_fuse}
    $\b{Pr}[p_i^M,\cdot] = \b{Pr}[p_i^M,\cdot] + \b{Pr}[p_j^M,\cdot]$\; \label{alg_quotient_up_s}
    Remove row $p_j^M$ from $\b{Pr}$\; \label{alg_quotient_up_fin_s}
    }\label{alg_quotient_end_loop}
\end{algorithm}

\textbf{Sub-Step 1.2. B\"uchi Petri net}. Starting from the B\"uchi automaton $B=\langle S,S_0,\Sigma_B,\rightarrow_B,F \rangle$ as in Def. \ref{def:Buchi}, Alg. \ref{alg:buchipn} obtains the corresponding B\"uchi Petri net system  $\mathcal{Q^B}$. For each state $s_i\in S$, a new place $p^B_i$ is added to the PN (line \ref{alg_buchi_places}). The first loop (lines \ref{alg_buchi_for_s} - \ref{alg_buchi_end_for_s}) is executed for each transition from a $s_i$ to a $s_j$ in the B\"uchi automaton. The second loop (lines \ref{alg_buchi_for_conj} - \ref{alg_buchi_end_for_conj}) is executed for each conjunctive element $\alpha_{k}$ from $\pi(s_i,s_j)$ and is adding a new transition $t_{\tau_k}$ in the B\"uchi Petri net from $p_i^B$ to $p_j^B$. Notice that all transitions corresponding to the conjunctive elements in $\pi(s_i,s_j)$ have the same input and the same output place from the B\"uchi PN. In line \ref{alg_buchi_set_m0}, the marking vector is initialized and it is updated in the loop in lines \ref{alg_buchi_for_m0} - \ref{alg_buchi_end_for_m0}. In particular, for each initial state of the B\"uchi automaton, one token is added in the corresponding place.

Alg. \ref{alg:buchipn} returns also \emph{virtual transitions} $T^V$, their relation with the final states being captured in $\b{Pre}^V$ and $\b{Post}^V$. These transitions will have zero firing cost when the solution of MILP \eqref{LPP1} is computed (explained in the next section) to maintain the B\"uchi PN in the final state if possible. One virtual transition is assigned to each final state $s_f$ of B\"uchi PN.

\begin{algorithm}[h]
\caption{B\"uchi Petri net}
\label{alg:buchipn}
\KwIn{$B=\langle S,S_0,\Sigma_B,\rightarrow_B,F \rangle$}
\KwOut{$\mathcal{Q}^B = \langle \langle P^B, T^B \cup T^V, [\b{Pre}^B\ \b{Pre}^V],$ $[\b{Post}^B\ \b{Post}^V] \rangle, \b{m}_0^B, {\cal Y}, h \rangle$}

Let $P^B = \{p^B_1, p^B_2, \ldots, p^B_{|S|}\}$ be the set of $|S|$ places\; \label{alg_buchi_places}
Let $T^B = \emptyset$ and $T^V = \emptyset$\;
\ForAll{$\left( s_{i}, \tau , s_j \right)\in \rightarrow_B$ \label{alg_buchi_for_s}}{ 
\ForAll{\text{conjuctive element} $\alpha_{k}$ \text{of} $\pi(s_i,s_j)$ \label{alg_buchi_for_conj}}{
    $T^B = T^B \cup t_{\tau_k}$ \tcc{add a new transition to $T^B$}
    Add a new column to $\b{Pre}^B$ and to $\b{Post}^B$ corresponding to $t_{\tau_k}$\;
    $\b{Pre}^B[p^B_i,t_{\tau_k}]=1$\;
    $\b{Post}^B[p^B_j,t_{\tau_k}]=1$\;
    }\label{alg_buchi_end_for_conj}
    }\label{alg_buchi_end_for_s}
Let $\b{m}_0^B = \b{0}^{|S|\times 1}$\; \label{alg_buchi_set_m0}
\ForAll{$s_i \in S_0$\label{alg_buchi_for_m0}}
{$\b{m}_0^B[p^B_i]=1$\; \label{alg_buchi_end_for_m0}}
\ForAll{$s_f \in F$\label{alg_buchi_for_f}}
{
$T^V = T^V \cup t_{s_f}$ \tcc{add a new \emph{virtual} transition to $T^V$}
    Add a new column to $\b{Pre}^V$ and to $\b{Post}^V$ corresponding to $t_{s_f}$\;
    $\b{Pre}^V[p^B_f,t_{s_f}]=1$\;
    $\b{Post}^V[p^B_f,t_{s_f}]=1$\;}
\end{algorithm}

\begin{example}\label{ex_buchiPN}
Let us consider the B\"uchi automaton in Fig.~\ref{fig:fig2}(a). By applying Alg.~\ref{alg:buchipn}, the B\"uchi PN in Fig.~\ref{fig:fig2}(b) is obtained. In this example, the places $p_2^B$ and $p_3^B$ matching states $s_2$ and $s_3$ are connected with a single transition corresponding to input $\pi(s_2,s_3) = y_1 \wedge y_2 \wedge y_3$. Near the transition labels it is written the input for B\"uchi that is required in order to fire the transition. Fig.~\ref{fig:fig2}(b) illustrates the addition of the virtual transition $t_1^V$ connected with the final state $s_3$ by a bi-directional arc. Once the system reaches this final state following the \textit{prefix}, the system persist in this state without additional cost in the proposed MILP. \hfill $\blacksquare$
\end{example}

\textbf{Sub-Step 1.3. Composition of Quotient RMPN and B\"uchi Petri net systems}. Alg. \ref{alg:rmpnreducedbuchipn} includes the full strategy to return a PN system $\mathcal{Q}^C$, by composition of the robotic team model ($\mathcal{Q}^M$ from Alg. \ref{alg:reduce_pn}) with the one of the specification ($\mathcal{Q}^B$ from Alg. \ref{alg:buchipn}). For this, a number of $2\times |\mathcal{Y}|$ places are needed, from which a half models the active observations (added in line \ref{alg_full_po}), while the other $|\mathcal{Y}|$ places model inactive observations (added in line \ref{alg_full_ipo}). Next line \ref{alg_full_mpo} puts zero tokens in the places modeling active observation places since we assume no active observation at the initial marking. In contrast, a number of tokens equal with the number of robots, i.e., $|R|$, are added to the inactive observations places. The sum of tokens in $p^{\neg O}_i$ and $p^O_i$ is always equal to $|R|$. In line \ref{alg_full_call_quotient}, Alg. \ref{alg:reduce_pn} is called to compute the $\mathcal{Q}^M$, while line \ref{alg_full_call_buchipn} call Alg. \ref{alg:buchipn} to compute the $\mathcal{Q}^B$. The next lines (\ref{alg_full_pos_full} - \ref{alg_full_trans_full}) define the sets of places, respectively transitions for the \textit{Composed PN} system $\mathcal{Q}^C$, while line \ref{alg_full_init_mark} defines the initial marking.


The matrices $\b{Pre}^C$ and $\b{Post}^C$ are initialized in lines \ref{alg_full_init_Pre} - \ref{alg_full_init_Post}. The last part of Alg. \ref{alg:rmpnreducedbuchipn} is responsible to combine the Quoting PN with B\"uchi Petri net by using the places modeling active and inactive observations. The loop in lines \ref{alg_full_for_connect_quotion_obs} to \ref{alg_full_endfor_connect_quotion_obs} is executing for each observation $y_i$. For each place $p_k$ of the Quoting PN  with output $y_i$, i.e., $y_i \in h(p_k)$, arcs from all input transitions to $p_k$ to place $p_i^O$ are added. Additionally, arcs from $p_i^O$ to all output transitions of $p_k$ are added as well. In this way, when a robot is entering in a region $p_k$ with output $y_i$, one token is added to $p_i^O$. Therefore, if $\b{m}[p_i^O] > 0 $ then observation $y_i$ is active. Place $p_i^{\neg O}$ is the complementary place of $p_i^O$ hence is connected with the same transitions as $p_i^O$ but with arcs oriented in the other sense. If $\b{m}[p_i^{\neg O}] = |R| $ then observation $y_i$ is not active.

Finally, loop in lines \ref{alg_full_for_connect_obs_buchi} to \ref{alg_full_endfor_connect_obs_buchi} connect the places modeling the active and inactive observations with the transitions of the B\"uchi PN. Since to each transition in B\"uchi $t_\tau^B$ a Boolean formula (a conjunction) is assigned, each atomic proposition that appears in the proposition is connected to places in sets $P^O$, respectively $P^{\neg O}$, in the following way. If the atomic proposition appears not negated, then is connected with a reading arc (a bidirectional arc) to the active observation place $p_i^O$ (line \ref{alg_full_not_negated_obs}). On the contrary, if the atomic proposition appears negated, then is connected with a reading arc of weight $|R|$ with the inactive observation place $p_i^{\neg O}$ (line \ref{alg_full_negated_obs}).

\begin{algorithm}[h]
\caption{Full PN system}
\label{alg:rmpnreducedbuchipn}
\KwIn{$\mathcal{Q} = \langle \langle P, T, \b{Pre}, \b{Post}\rangle, \b{m}_0, {\cal Y}, h \rangle$, $B=\langle S,S_0,\Sigma_B,\rightarrow_B,F \rangle$}
\KwOut{$\mathcal{Q}^C = \langle \langle P^C, T^C, \b{Pre}^C, \b{Post}^C \rangle, \b{m}_0^C, {\cal Y}, h \rangle$}

Let $P^O = \{p^O_1, p^O_2, \ldots, p^O_{|\mathcal{Y}|}\}$ be the set of $|\mathcal{Y}|$ places modeling the active observations\; \label{alg_full_po}
Let $P^{\neg O} = \{p^{\neg O}_1, p^{\neg O}_1, \ldots, p^{\neg O}_{|\mathcal{Y}|}\}$ be the set of $|\mathcal{Y}|$ places modeling the inactive observations\; \label{alg_full_ipo}
Let $\b{m}_0^O = \left[ \b{0}^{|P^O|\times 1},\ |R|\cdot \b{1}^{|P^{\neg O}|\times 1} \right]$\;\label{alg_full_mpo}
Execute Alg. \ref{alg:reduce_pn} and let $\mathcal{Q}^M = \langle \langle P^M, T^M, \b{Pre}^M, \b{Post}^M \rangle, \b{m}^M_0, {\cal Y}, h \rangle$ be the returned PN\; \label{alg_full_call_quotient}
Execute Alg. \ref{alg:buchipn} and let $\mathcal{Q}^B = \langle \langle P^B, T^B \cup T^V,$ $[\b{Pre}^B\ \b{Pre}^V], [\b{Post}^B\ \b{Post}^V] \rangle, \b{m}_0^B, {\cal Y}, h \rangle$ be the returned PN\; \label{alg_full_call_buchipn}
Let $P^C = P^M \cup P^B \cup P^O \cup P^{\neg O}$\; \label{alg_full_pos_full}
Let $T^C = T^M \cup T^B \cup T^V$\; \label{alg_full_trans_full}
$\b{m_0^C} = \left[ \b{m}^M_0,\ \b{m}_0^B, \b{m}_0^O\right]$\;\label{alg_full_init_mark}
Let $\b{Pre}^C=\left[ \begin{array}{ccc}
    \b{Pre}^M & \b{0}^{|P^M|\times|T^B|} & \b{0}^{|P^M|\times|T^V|}\\
    \b{0}^{|P^B|\times|T^M|} & \b{Pre}^B & \b{Pre}^V\\
    \b{0}^{|P^O|\times|T^M|} & \b{0}^{|P^O|\times|T^B|} & \b{0}^{|P^O|\times|T^V|}\\
    \b{0}^{|P^{\neg O}|\times|T^M|} & \b{0}^{|P^{\neg O}|\times|T^B|} & \b{0}^{|P^B|\times|T^V|}
\end{array} \right]$\; \label{alg_full_init_Pre}
Let $\b{Post}^C=\left[ \begin{array}{ccc}
    \b{Post}^M & \b{0}^{|P^M|\times|T^B|} & \b{0}^{|P^M|\times|T^V|}\\
    \b{0}^{|P^B|\times|T^M|} & \b{Post}^B & \b{Post}^V\\
    \b{0}^{|P^O|\times|T^M|} & \b{0}^{|P^O|\times|T^B|} & \b{0}^{|P^O|\times|T^V|}\\
    \b{0}^{|P^{\neg O}|\times|T^M|} & \b{0}^{|P^{\neg O}|\times|T^B|} & \b{0}^{|P^O|\times|T^V|}
\end{array} \right]$\; \label{alg_full_init_Post}
\ForAll{$y_i \in \mathcal{Y}$\label{alg_full_for_connect_quotion_obs}}{
Let $P'=\{p \in P^M | y_i \in h(p)\}$\;
\ForAll{$p_k \in P'$}{
$\b{Post}^C[p_i^O, \preset{p_k}]=\b{Pre}^C[p_i^O, \postset{p_k}]=1$\;
$\b{Pre}^C[p_i^{\neg O}, \preset{p_k}]=\b{Post}^C[p_i^{\neg O}, \postset{p_k}]=1$\;
}
}\label{alg_full_endfor_connect_quotion_obs}

    \ForAll{$t_\tau^B \in T^B$\label{alg_full_for_connect_obs_buchi}}{
    Let $\pi_i$ be the DNF formula assigned to $t_\tau^B$\;
    \If{$\pi_i \neq \top$}{
    \ForAll{atomic propositions $y_i$ appearing \textbf{not negated} in $\pi_i$}{\label{alg_full_not_negated_obs}
        $\b{Pre}^C[p^O_i,t_\tau^B]=\b{Post}^C[p^O_i,t_\tau^B]=1$\;}
    \ForAll{atomic propositions $y_i$ appearing \textbf{negated} in $\pi_i$}{\label{alg_full_negated_obs}
        $\b{Pre}^C[p^{\neg O}_i,t_\tau^B]=\b{Post}^C[p^{\neg O}_i,t_\tau^B]=|R|$\;}
    }
    }\label{alg_full_endfor_connect_obs_buchi}

\end{algorithm}


Following the Alg. \ref{alg:rmpnreducedbuchipn}, note that transitions $T^B$ can be fired if a set of observations are active and some are not. For example, transition $t_1^B$ in PN of Fig.~\ref{fig:fig2}(b) can be executed if $y_1$ and $y_2$ are active. For this reason, transition $t_1^B$ is connected by self-loops with places $p^o_1$ and $p^o_2$. Thus, $t_1^B$ can be fired only if the corresponding observations are active. The fired transition leads to a new marking in $\mathcal{Q}^C$ RMPN system.

\begin{figure*}
    \centering
 \includegraphics[width=.55\textwidth]{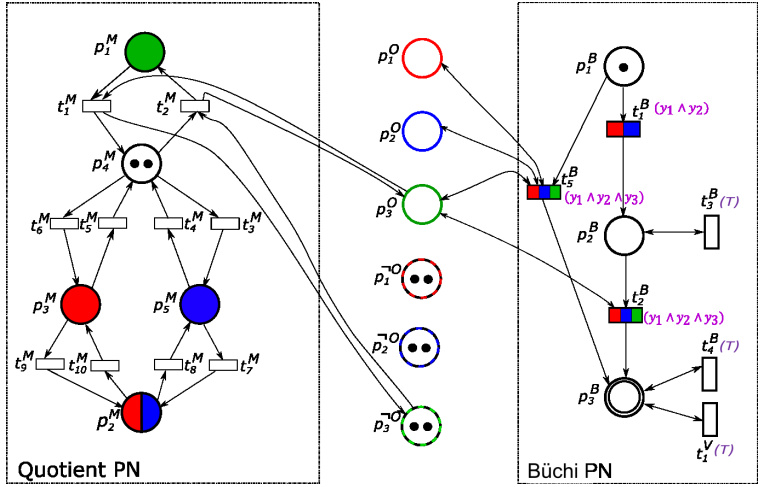}
    \caption{Partially Composed Petri net, based on active and inactive observations of $y_3$}
    \label{fig:composed_pn}
\end{figure*}

Fig. \ref{fig:composed_pn} depicts a part of \textit{Composed Petri net} with its initial marking, returned by Alg. \ref{alg:rmpnreducedbuchipn}. Quotient PN $\mathcal{Q}^M$ and B\"uchi PN $\mathcal{Q}^B$ are linked with the places representing the active and inactive observation $y_3$. For the sake of clarity, Fig. \ref{fig:composed_pn} considers only the arcs for one region on interest ($y_3$). As explained in Alg. \ref{alg:rmpnreducedbuchipn}, when one transition in $\mathcal{Q}^M$ is fired and the observation is changed, the reduced PN of the environment deposit one token to the respective active observation, e.g., $t_2^M$ is enabled when $y_3$ is not active (here the robots being in the free space initially), and if it fires a token to both $p_1^M$ and $p_3^O$ are produced. In $\mathcal{Q}^B$, the transitions are fired based on the assigned Boolean formula. Herein, the places for active and inactive observations are required to enable these transitions, e.g., $t_5^B$ and $t_2^B$ depend on active observation $y_3$, as it is illustrated in Fig. \ref{fig:composed_pn}. The transitions in B\"uchi PN are colored according to the required active observations, i.e., red - $y_1$, blue - $y_2$, green - $y_3$. The rationale for the rest of connections is maintained for the active and inactive observations modeling $y_1$ and $y_2$. 

\textbf{Complexity.} The algorithms from the first part of the solution (designing the \textit{Composed PN}) are tractable, being based on simple operations. Based on each algorithm's inputs, the complexity is as follows: (a) Alg. \ref{alg:reduce_pn} is polynomial ($\mathcal{O}(n^2)$) depending on the number of places ($n = |P|$), subject to the environment decomposition technique; (b) Alg. \ref{alg:buchipn} is polynomial with respect to the inputs $\rightarrow_B$ of the B\"uchi automaton $\mathcal{B}$ and the number of atomic propositions over set $\mathcal{Y}$. In general, the LTL formulas specified for multi-robot systems have a reduced number of atomic propositions, the exponential upper-bound of $2^{|\mathcal{Y}|}$ being seldom achieved. Lastly, (c) Alg. \ref{alg:rmpnreducedbuchipn} is polynomial over cardinality of set $\mathcal{Y}$ and of set of B\"uchi's transitions $T^B$.




\section{Complete solution}\label{sec:project}
This section is dedicated to return a solution in terms of robot's paths to ensure the given LTL mission. In this sense, two actions are necessary. First, a solution is obtained based on the composed PN model $\mathcal{Q}^C$ computed previously, which captures the evolution of robots with respect to the LTL mission. Secondly, the solution is projected into the original RMPN model, outputting robots movement towards their goal positions.

\textbf{Sub-Step 2.1. Solution on the reduced model}. The main idea of MILP \eqref{LPP1} is to drive the PN to a state corresponding to a final state in B\"uchi (marking $\b{m}_{2k}^C$). The solution of this MILP is divided into 2 parts, \textit{prefix} and \textit{suffix}, each based on the parameter $k$. The motivation behind the division lies in minimizing the run time for finding the full solution, since, in general, it is faster to solve two small MILP problems than a big one. The difference between these MILPs dwells in the initial marking $m_0^C$ of \textit{Composed Petri net} model: represented by the initial state for \textit{prefix} and by the previously reached final state of B\"uchi for \textit{suffix}.

The MILP \eqref{LPP1} considers $k$ steps for {\it prefix}, respectively {\it suffix}. Therefore, $k$ is a design parameter that will correspond to the maximum number of states in \textit{prefix} and \textit{suffix} of accepted runs in B\"uchi automaton $B$, where $k \geq 1$. For each odd step, a transition in Quotient PN is fired, while for each even step, a transition in B\"uchi PN is fired. Because to create a token in a place of B\"uchi PN it may be necessary to move a robot through all places of the Quotient PN, an upper-bound of parameter $k$ can be imposed, as follows: $\left(|P^M|-1\right) \times \left(|P^B|-1\right)$. If parameter $k$ is chosen very small, a marking corresponding to a final state in B\"uchi PN cannot be reached being necessary in this case to increase $k$. 





Explanations on MILP \eqref{LPP1} are as follows:
\begin{itemize}
    \item The cost function minimizes the number of fired transitions of Quotient and B\"uchi PNs, scaled by index $i$, to enforce reaching the solution during the first steps of the defined horizon of $k$ steps, when it is possible.
    \item Constraints (3b) correspond to the state equation \eqref{eq:steq}.
    \item Set of constraints (3c) allow robots to advance to only one place in Quotient PN. Note that by firing a transition in the Quotient PN, the observations are changed (or at least the regions that generate the observations) such that, in the following step (when $i$ is even) one transition in B\"uchi should be fired.
    \item Set of constraints (3d) imposes that the first transition to fire from the B\"uchi PN to be different by a virtual one, thus force the B\"uchi to leave the initial state. 
    \item Set of constraints (3e) allow the firing only of one transition corresponding to the B\"uchi PN. Notice that here, a virtual transition that has zero cost can fire also such that if the prefix or suffix is reached in a smaller steps than $k$, no other transitions will be executed.
    \item Constraints (3f) ensures the marking after $k$ steps is the final state in B\"uchi, denoted as $p_f^B$ being an input parameter.
\end{itemize}

\textbf{Parameters:}\\
\begin{tabular}{rl}
$|R|$ &  - number of robots \\
$p_f^B$ & - place modeling a final state in B\"uchi \\
$\b{C}^C$ & - token flow matrix of the Complete PN \\
$\b{Pre}^C$ & - pre-incidence matrix of the Complete PN\\
\end{tabular}\\

\textbf{Variables:}
\begin{itemize}
    \item $\b{m}_i^C = \left[ \b{m}_i^M\ \b{m}_i^B\ \b{m}_i^O\ \b{m}_i^{\neg O}\right]$ - marking at step $i$ of the full PN composed by the marking of Quotient PN ($\b{m}_i^M$), B\"uchi PN ($\b{m}_i^B$), active observation places ($\b{m}_i^O$) and inactive observation places ($\b{m}_i^{\neg O}$);
\item $\b{\sigma}_i^C = \left[ \begin{array}{c}\b{\sigma}_i^M\\ \b{\sigma}_i^B\\ \b{\sigma}_i^V \end{array} \right]$ - firing vector at step $i$ of the full PN, composed by the firing vector of Quotient PN ($\b{\sigma}_i^M$), B\"uchi PN ($\b{\sigma}_i^B$) and of virtual transitions ($\b{\sigma}_i^V$). 
\end{itemize}
\textbf{Objective:}
\begin{subequations}\label{LPP1}
\begin{align}
\min & \sum \limits_{i=1}^{2 \cdot k} i \cdot  \left( \b{1}^T \cdot \b{\sigma}_i^M + \b{1}^T \cdot \b{\sigma}_i^B \right)
\end{align}

\textbf{Constraints:}
\begin{align}
 \b{m}_i^C - \b{m}_{i-1}^C - \b{C}^C \cdot \b{\sigma}_i^C & = 0, && \scriptstyle i=\overline{1, 2\cdot k} \\
  \begin{array}{r}
  \b{m}_i^C - \b{Pre}^C\cdot \b{\sigma}_i^C \\ 
  \b{1}^T \cdot \b{\sigma}^M_i \\
  \b{1}^T \cdot \b{\sigma}^B_i +
  \b{1}^T \cdot \b{\sigma}^V_i \end{array}  & \left. \begin{array}{l}
  \geq 0, \\
  \leq |R|, \\
  = 0,
  \end{array} \right\}
  && \begin{array}{l} \scriptstyle i=2\cdot j + 1 \\ \scriptstyle j = \overline{0, k-1} \end{array} \\
  \begin{array}{r}
  \b{1}^T \cdot \b{\sigma}^M_i \\
  \b{1}^T \cdot \b{\sigma}^B_i \\
  \b{1}^T \cdot \b{\sigma}^V_i \end{array}  & \left. \begin{array}{l}
  = 0, \\
  = 1, \\
  = 0,
  \end{array} \right\}
  && \scriptstyle  i=2  \\
    \begin{array}{r}
  \b{1}^T \cdot \b{\sigma}^M_i \\
  \b{1}^T \cdot \b{\sigma}^B_i +
  \b{1}^T \cdot \b{\sigma}^V_i \end{array}  & \left. \begin{array}{l}
  = 0, \\
  = 1, \\
  \end{array} \right\}
  && \begin{array}{l} \scriptstyle i=2\cdot j \\ \scriptstyle j = \overline{2, k} \end{array} \\
  \b{m}_i^C[p_f^B] & = 1, && \scriptstyle i=2\cdot k
\end{align}
\end{subequations}

\textbf{Sub-Step 2.2. Projecting the solution}. First, from the complete solution of MILPs \eqref{LPP1} it is extracted the sequence of markings corresponding to the robot team in the Quotient PN $\mathcal{Q}^M$. Let $M=\langle \b{m}_1^M, \b{m}_2^M, \ldots, \b{m}_{2k}^M\rangle$ be that sequence. Before projecting the solution, the identical marking from $M$ are removed. In particular, if $\exists \ \b{m}_i^M, \b{m}_{i+1}^M \in M$ such that $\b{m}_i^M = \b{m}_{i+1}^M$, then $\b{m}_i^M$ is removed from $M$. The number of markings in $M$ is, in general, reduced. Notice that only successive identical markings are removed while the marking which register the end of the \textit{prefix} and beginning of \textit{suffix} is stored. Additionally, let us add $\b{m}_0^M$ as the first element in $M$. 

Furthermore, let $G = \langle \b{g}_1, \b{g}_2, \ldots, \b{g}_{2k} \rangle$ be a sequence of $2k$ vectors such that, $\b{g}_i \in \{0,1\}^{|P^M|}$ and $\b{g}_i[j] = 1$ if $\b{m}_i^M[j]=0$, and $\b{g}_i[j] = 0$ otherwise. This vector will be used in MILP \eqref{LPP_projecting} to project the solution and to not allow to activate other observation between two steps.

The following MILP extends every marking from the Quotient RMPN into a string of markings in the original RMPN. The string has the same active observations, to ensure the validity of the LTL formula when exist finite repetitions. In order to avoid collisions, a number of $|R|$ intermediate markings are additionally introduced between two successive marking, to ensure that each region of the original net is crossed by maximum one robot. Assuming that at the initial state each region contains maximum one robot, following MILP ensures the collisions free trajectories, although the active observations are maintained.

\textbf{Parameters:}\\
\begin{tabular}{rl}
$|R|$ & - number of robots\\
$M$ & - sequence of markings returned by \eqref{LPP1} \\
$\b{Pr}$ & - projection matrix between $Q^B$ and RMPN \\
$\b{C}$ & - token flow matrix of the RMPN\\
$\b{Pre}, \b{Post}$ & - pre/post-incidence matrices of RMPN\\
$\b{m}_i^M$ & - marking at step $i$ of the Quotient PN.
\end{tabular}

\textbf{Variables:}\\
\begin{itemize}
    \item $\b{m}_{i,j}$ - marking at step $i$ of RMPN, considering the intermediate marking $j$ 
    \item $\b{\sigma}_{i,j}$ -  firing vector at step $i$ of RMPN, for the intermediate marking $j$ with $i=\overline{0,|M|} ,  j=\overline{1,|R|+1} $
\end{itemize}

\textbf{Objective:}
\begin{subequations}\label{LPP_projecting}
\begin{align}
\min \b{1}^T\cdot \sum_{i,j} \b{\sigma}_{i,j}
\end{align}

\textbf{Constraints:}
\begin{align}
  \b{m}_{i,j} - \b{m}_{i,j-1} - \b{C} \cdot \b{\sigma}_i & = \b{0}, &  \scriptstyle i=\overline{0,|M|} ,  j=\overline{1,|R|+1}\\
  \b{m}_{i,0} - \b{m}_{i-1,|R|+1}  & = \b{0}, & \scriptstyle i=\overline{1,|M|} \\
  \b{Pr} \cdot \b{m}_{i,j} - \b{m}_i^M & = \b{0}, & \scriptstyle i=\overline{0,|M|} , j=\overline{1,|R|}\\
  \b{Post}[\b{g}_i \cdot \b{Pr},\cdot] \cdot \b{\sigma}_{i,j} & =\b{0}, & \scriptstyle i=\overline{0,|M|} ,  j=\overline{1,|R|+1}\\
  \b{Post} \cdot \b{\sigma}_{i,j} + \b{m}_{i,j-1} & \leq \b{1}, & \scriptstyle i=\overline{0,|M|} ,  j=\overline{1,|R|+1}\\
  \b{m}_{i,|R|+1} - \b{Pre} \cdot \b{\sigma}_{i,|R|+1} & \geq \b{0}, & \scriptstyle i=\overline{0,|M|}
\end{align}
\end{subequations}

The constraints in MILP \eqref{LPP_projecting} are the followings:
\begin{itemize}
    \item (4b) is the state equation \eqref{eq:steq} of the RMPN model; 
    \item (4c) ensure that the last intermediate marking is the same with the initial marking in the next step; \item (4d) keep the same observations during the $\b{m}_{i,1}$ to $\b{m}_{i,|R|}$. In fact, these $|R|$ intermediate markings are introduced in order to avoid collisions between $\b{m}_i^M$ and $\b{m}_{i+1}^M$.
    \item Constraints (4e) ensure that the corresponding firing vectors of the intermediate markings from $\b{m}_{i,1}$ to $\b{m}_{i,|R|}$ should not activate other observations; 
    \item (4f) ensure the collision avoidance between successive markings of the original RMPN by imposing that the number of times that each region is crossed between two intermediate markings is maximum one; 
    \item Last constraints, (4g) impose that the movements of the robots from sub-step $\b{m}_{i,|R|}$ to sub-step $\b{m}_{i,|R|+1}$ is done synchronously by all robots (each robot fires only one transition) such that the generated observation of $\mathcal{Q}^M$ changes according to the transitions fired in B\"uchi. 
\end{itemize}

\textbf{Global algorithm.} Alg. \ref{alg:findingsol} is responsible to find a complete solution for the robot trajectories, that fulfills the given mission $\varphi$. For this reason, several inputs are required as it can be seen in Alg. \ref{alg:findingsol}. They will be used to (i) compute a solution in the reduced model, and to (ii) project the previous solution in the original RMPN model.

The lines 3-14 search for a feasible $Run$ in \textit{Composed PN} based on MILP \eqref{LPP1}, for all places modeling the final states in B\"uchi (captured in set $Set_f$). As a first step, the \textit{prefix} is computed (line \ref{alg_complete_prefix}). If the \textit{prefix} exists (line \ref{alg_complete_start_pref}), then the \textit{suffix} is computed. Let us recall that one refers to an active observation if the place $p_i^O$ assigned to observation $y_i$ has at least one token. Line \ref{alg_complete_start_suff} evaluates if the last active observations (denoted with $P^O_f$) are included in the self-loop of the final state $s_f$. If this condition in line \ref{alg_complete_start_suff} is respected, the \textit{suffix} is returned by MILP \eqref{LPP1}. Otherwise, the \textit{suffix} is represented by the final state, without being necessary to solve the MILP \eqref{LPP1} (line \ref{alg_complete_end_pref}). For both \textit{prefix} and \textit{suffix} not empty, the feasible \textit{Run} is computed in line \ref{alg_complete_run} which should ensure the LTL mission. Based on this \textit{Run}, the robots trajectories are projected into the original RMPN $\mathcal{Q}$ (line \ref{alg_complete_computetraj}). If no feasible $Run$ was computed for any final state in B\"uchi (line 19), then parameter $k$ is increased and the procedure from lines 3-14 is performed.

In case $k$ reaches its upper-bound and no feasible $Run$ could be computed, the Alg. \ref{alg:findingsol} cannot return a solution. This means that the problem does not have a solution with respect to the team of robots and the LTL mission, e.g., three disjoint regions should be visited in the same time, but the team contains two robots. This algorithm guarantees to obtain a solution if at least one exists (lines 3-14), which is correct both in terms of team capabilities and LTL satisfaction. The upper-bound of $k$ ensure that a final state of B\"uchi is reached by prefix and revisited by suffix whenever possible.

\begin{algorithm}
\caption{Global solution for robot's trajectories}
\label{alg:findingsol}
	\KwIn{RMPN $\mathcal{Q}$, \textit{Composed Petri net} $\mathcal{Q}^C$, set $\mathcal{Y}$, $|R|$, set of active observations $P^O$, set of final states $F \in B$ , finite horizon $k$}
	\KwOut{Robot movement strategies}
	Let $Set_f = \{p_{f_1}^B, p_{f_2}^B, \dots p_{f_n}^B\}$\;
	Let Run = $\emptyset$\;
	\While{$Set_f \neq \emptyset$ }{
	Select place $p_{f_i}^B$ modeling a final state $s_f$ in B\"uchi\;
    Compute \textit{prefix} for $p_f^B$ based on MILP \eqref{LPP1}\;\label{alg_complete_prefix}
    \If{\emph{prefix} $\neq \emptyset$}{ \label{alg_complete_start_pref}
    	    Let $P^O_f$ be the last active observations for $p_f^B$\;
    	 \uIf{$P^O_f \cancel{\models} \pi(s_f,s_f) $}{ \label{alg_complete_start_suff}
    	 Let $\b{m}_0^C = \b{m}_{k}^C$, where $\b{m}_{k}^C$ is the solution of MILP \eqref{LPP1}\;
            Compute \emph{suffix} based on MILP \eqref{LPP1}\;
            } \label{alg_complete_end_suff}
	    \Else
	        {\emph{suffix} = $s_f$\;\label{alg_complete_end_pref}}
	}
    \If{(\emph{prefix} $\neq \emptyset$ AND \emph{suffix} $\neq \emptyset$)}{
    \emph{Run} = \textit{prefix suffix suffix \dots}\; \label{alg_complete_run}}}
	\If{Run $\neq \emptyset$}{
	Compute \emph{Robot movement strategies} based on MILP \eqref{LPP_projecting} for the returned $Run$\; \label{alg_complete_computetraj}}
	\ElseIf{$k < \left( \left(|P^M|-1\right)\times \left(|P^B|-1\right) \right)$}
    {Increase parameter $k$\;
    Compute Run for new $k$ \tcc{lines 3-14}}
	\Else{Return false. The formula cannot be achieved by the robots in the actual environment.}

\end{algorithm}

It is worth mentioning that our approach has a slower increase of complexity than other methods which are based on a product of automatons. Herein, the maximum number of places in the \textit{Composed Petri net} $\mathcal{Q}^C$ is given by the sum of places for $\mathcal{Q}^M$, the number of places for $\mathcal{Q}^B$ and twice the number of ROIs, $2\cdot|\mathcal{Y}|$. Thus, the state explosion problem is avoided. Although optimality is not achieved in terms of trajectories length or number of firings in the original RMPN for a single repetition of suffix, the solution obtained in the reduced model can be improved by considering all the final states $s_f$ in Alg. \ref{alg:findingsol} while choosing a desired run rather than stopping at the first obtained one.

\textbf{Complexity.} The complexity of this algorithm is NP-hard, justified by the use of MILP problems. The total number of unknown variables in both MILPs is given by the number of markings and transitions in $\mathcal{Q}^C$, respectively the number of markings and transitions in $\mathcal{Q}$. On the other hand, the total number of constraints is related to the number of intermediate markings for MILP \eqref{LPP1}, while MILP \eqref{LPP_projecting} depends on the solution of the previous MILP.

\section{Simulation results and comparison to \cite{kloetzer2020path}}\label{sec:comp_simul}

The solution proposed in this paper was implemented and integrated in RMTool - MATLAB \cite{IPGoMaKl15}, using the CPLEX Optimizer \cite{CPLEX} solver for the mentioned MILPs from the previous section. The results captured in this section were obtained using a laptop with i7 - $8^{th}$ gen. CPU @ 2.20GHz and 8GB RAM. 

\begin{example}\label{ex_easyexample}
Let us recall the environment from Fig. \ref{fig:fig1} (a) and the LTL mission that needs to be fulfilled, as in equation \eqref{eq:1}: $$\varphi= \diamondsuit \left( y_1 \wedge y_2 \wedge y_3 \right) \wedge \neg \left( y_1 \vee y_2 \right) \mathcal{U} \left( y_1 \wedge y_2 \right).$$ 

The robots need to reach all three regions of interest at the same time, ensuring $y_1$ and $y_2$ are reached simultaneously. Figure \ref{fig:fig1}(a) illustrates the robot trajectories returned by Alg. \ref{alg:findingsol}. It can be observed that both robots moves towards the regions $y_1$ and $y_2$ to complete the concurrently requirement. Afterwards, one of the robots ($r_1$) advances and enters the last region of interest $y_3$, while the second robot enters $p_{13}$ = \{$y_1, y_2$\} in the same time. Several numerical results can be specified, based on the described steps in Section \ref{sec_solution}, such as: the resulting \emph{Quotient PN model} with 5 places and 10 transitions computed in 0.02 seconds, the \emph{B\"uchi automaton} with 3 states returned in 0.16 seconds and the \emph{Composed PN} model with 14 places and 16 transitions (in account of the added virtual transition) computed in 0.02 seconds. MILP \eqref{LPP1} reaches the final state in B\"uchi in 0.05 seconds, using 180 unknown variables (for $k$ = 6). The MILP \eqref{LPP1} returns the \textit{prefix} $s_1s_2$, without the necessity of solving MILP \eqref{LPP1} to compute the \textit{suffix}, because the final state $s_3$ is True $\top$ for its self-loop. The last phase of the complete algorithm lies in projecting the solution returned so far in the original RMPN, based on MILP \eqref{LPP_projecting}. The run time is 0.05 seconds for 900 unknown variables, with the cost function equal with 11 (minimum number of cells crossed by the robots).

Notice that with the inclusion of virtual transitions connected with the final states, the minimum value of MILP \eqref{LPP1} is independent of the value of $k$ parameter.\hfill $\blacksquare$
\end{example} 

\begin{figure*}
    \centering
 \includegraphics[width=1.0\textwidth, height = 140pt]{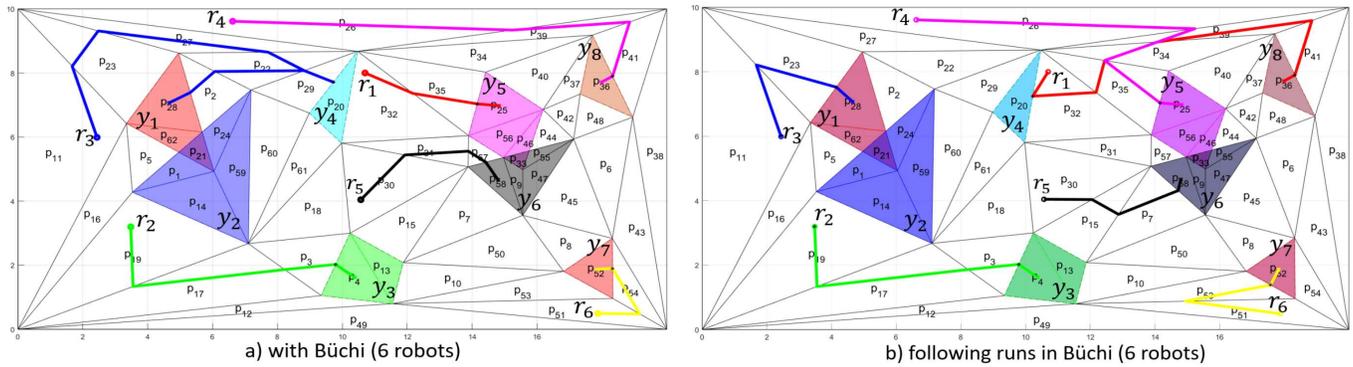}
    \caption{Trajectories satisfying the LTL formula from Example \ref{ex_w_folBuchi} (red - $r_1$; green - $r_2$, blue - $r_3$, magenta - $r_4$, black - $r_5$, yellow - $r_6$)}
    \label{fig:ex_w_folBuchi}
\end{figure*}

The result analysis is based on a comparison of the current approach from Alg. \ref{alg:findingsol} - joined model of \textit{Composed Petri net} model,  in contrast with the sequential procedure of the two models PN and B\"uchi automaton, captured in \cite{kloetzer2020path}. We will refer to the first approach as \textit{with B\"uchi}, while the second approach will be denoted as \textit{following B\"uchi}. The comparison does not include methods based on discrete transition systems by cause of the state explosion problem. As a reminder, the method from \cite{kloetzer2020path} consist in a iteration of a finite number of acceptable runs in B\"uchi automaton which satisfy the LTL formula. Then it examines the sequence of reachable markings for which the desired observations can be activated, while respecting the desired run. Both works acknowledge the benefits of B\"uchi automaton model and Petri net models, while maintaining the collision avoidance between the robots. Nevertheless, the contribution of the current method prevail a complete algorithm, based on the joined model $\mathcal{Q}^C$. It should be noted that the collision avoidance problem is solved in \cite{kloetzer2020path} only when several restrictions make good use of a parameter with a suitable chosen value. Otherwise, the collision free trajectories cannot be guaranteed. On the other hand, the current approach solves this problem based only on the set of restrictions (4f) of MILP \eqref{LPP_projecting}.

\begin{example}\label{ex_w_folBuchi}
We consider the next LTL formula, by maintaining the synchronization operation illustrated previously: $\varphi= \square \left(\diamondsuit y_1 \wedge \diamondsuit y_3 \wedge \diamondsuit y_5 \wedge \diamondsuit y_6 \wedge \diamondsuit y_7 \wedge \diamondsuit y_8 \right) \wedge \neg \left( y_5 \vee y_6 \right) \mathcal{U} \left( y_5 \wedge y_6 \right) \wedge \neg \left( y_4 \vee y_7 \right) \mathcal{U} \left( y_4 \wedge y_7 \right).$ This specification imposes the visit of several ROIs in an environment with 8 regions of interest, while the regions $y_5$ and $y_6$ are simultaneously reached, respectively $y_4$ and $y_7$. Table \ref{table:comparison} captures the comparison between the mentioned methods in terms of running time and value of the cost function (number of crossed cells following the robot trajectories). The table accounts scenarios with 4, 5, 6 and 10 robots, and Fig. \ref{fig:ex_w_folBuchi} exemplifies robots trajectories for 6 robots. The synchronization points for the entire team (black stars) are depicted along the trajectories right before entering a regions of interest. Both approaches consider an equal the number of intermediate markings ($k = 10$). 

\begin{table}[h!]
\centering
\caption{Comparison between current approach \textit{with B\"uchi} and \textit{following B\"uchi} captured in \cite{kloetzer2020path}.}
\begin{tabular}{||c| c c | c c ||} 
 \hline
 \multirow{5}{4em}{Number of robots} & \multicolumn{2}{c}{\multirow{2}{10em}{\textbf{Run time to return a solution [sec]}}} & \multicolumn{2}{|c||}{\multirow{2}{10em}{\textbf{Cost function value}}} \\ [4ex] 
 \multirow{4}{4em}{} & \multirow{2}{5em}{\textit{following B\"uchi}} & \multirow{2}{5em}{\textit{with B\"uchi}} & \multirow{2}{5em}{\textit{following B\"uchi}} & \multirow{2}{5em}{\textit{with B\"uchi}} \\ [2.5ex]
 \hline\hline
\multirow{1}{4em}{4} & \multirow{1}{5em}{9.56} & \multirow{1}{5em}{0.75} & \multirow{1}{5em}{37} & \multirow{1}{5em}{39}\\ [0.3ex] 
\hline
\multirow{1}{4em}{5} & \multirow{1}{5em}{2.22} & \multirow{1}{5em}{0.26} & \multirow{1}{5em}{23} & \multirow{1}{5em}{28}\\ [0.3ex] 
\hline
\multirow{1}{4em}{6} & \multirow{1}{5em}{0.77} & \multirow{1}{5em}{0.11} & \multirow{1}{5em}{20} & \multirow{1}{5em}{21}\\ [0.3ex] 
\hline
\multirow{1}{4em}{10} & \multirow{1}{5em}{1.2} & \multirow{1}{5em}{0.78} & \multirow{1}{5em}{13} & \multirow{1}{5em}{13}\\ [0.3ex] 
 \hline 
\end{tabular}
\label{table:comparison}
\end{table}

The procedure \textit{with B\"uchi} returns the solution for a RMPN model with 37 places and 67 transitions, and the procedure \textit{following B\"uchi} computes the solution for RMPN model with 62 places and 182 transitions. The running time for the current method is smaller than the previous approach, as a result of solution division in \textit{prefix} and \textit{suffix}. For the example illustrated in Fig.~\ref{fig:ex_w_folBuchi}, the MILP \eqref{LPP1} assigned to \textit{suffix} is not computed, by cause of last active observations being included in the self-loop of the final state in B\"uchi.\hfill $\blacksquare$
\end{example}

Let us recall the meaning of the cost function for MILP \eqref{LPP_projecting} as being the total number of fired transitions in the original RMPN model. Although in the table it can be seen that the current approach returns a cost function value bigger or equal compared with the method from \cite{kloetzer2020path}, this fact is not always true. Let us consider the counterexample with the LTL formula: $\varphi= \diamondsuit y_2 \wedge \square \diamondsuit \left( y_1 \wedge \diamondsuit y_3 \right) \wedge \neg y_3 \mathcal{U} y_2$ applied for the same environment from Fig. \ref{fig:fig1}. Informally, this requirement specifies the visit of region $y_2$ before $y_3$ sometime along the trajectory and the visit of $y_1$ and $y_3$ infinitely often. In this sense, a certain order for visiting the ROIs is imposed. After running the Alg. \ref{alg:findingsol}, the complete solution is obtained 0.05 seconds by solving both MILPs (\ref{LPP1} and \ref{LPP_projecting}) for the joined model $\mathcal{Q}^C$ with 19 places and 36 transitions. The B\"uchi automaton of the LTL specification contains two final states, for each of those returning the cost function 6, respectively 10. The current method projects the solution for the final state with the smaller cost function (6), while the method \textit{following B\"uchi} returns the cost function equal with 10.

\section{Conclusion}\label{sec:concl}

This paper proposes a complete algorithm in terms of motion planning for dynamic multi-robot systems. A global LTL specification is given for a team of identical robots, which should reach and/or avoid several regions of interest in the environment. Two formalisms are used for the collision free trajectories: (1) Robot Motion Petri net (RMPN) model for to the environment, and (2) B\"uchi automaton for the LTL formula. The proposed algorithm values the advantages of both models by a joint new representation denoted \textit{Composed Petri net} system. 


The efficiency of both the proposed \textit{Composed Petri net} model and the complete algorithm were illustrated in the numerical evaluation, considering a comparison with a previous work \cite{kloetzer2020path}. The current method offers a complete solution based on the union of both models (environment and mission), in contrast with the previous work based on iterative and sequential approach of the models which cannot assure the completeness in terms of robot trajectories. In addition, the method proves to increase slower in terms of complexity (run time) compared with techniques based on automaton products, in account of a smaller number of places in the \textit{Composed PN}.

The future work envisions to further reduce the complexity, while considering partial unknown environment. One approach in this sense is based on distributive algorithms in comparison with the centralized current method, e.g., each robot computes its own trajectory. 

\bibliographystyle{IEEEtran}
\bibliography{references}

\begin{IEEEbiography}[{\includegraphics[width=1in,height=1.25in,clip,keepaspectratio]{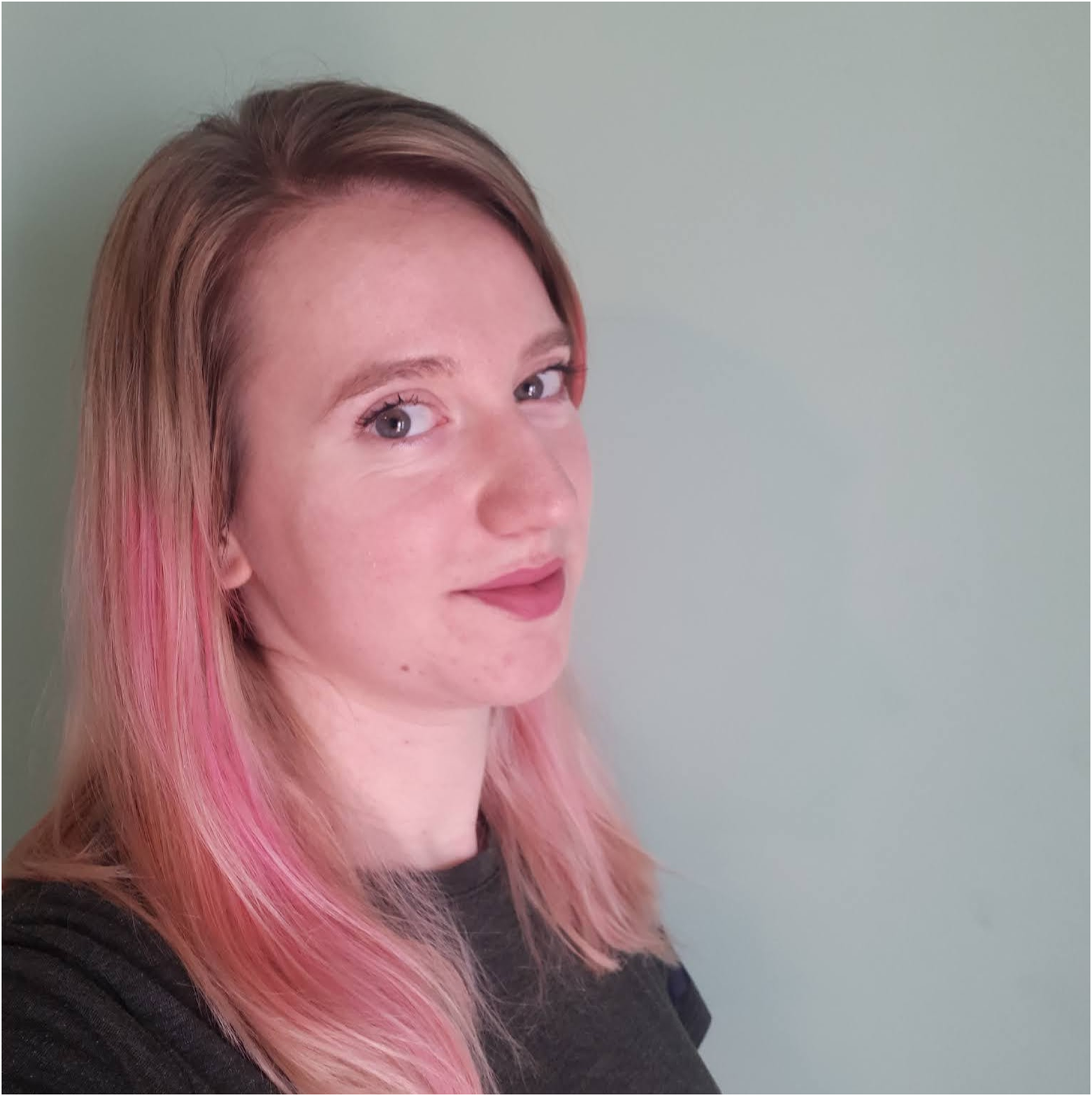}}]{Sofia Hustiu} received the B.S. and M.Sc. degrees in automatic control and applied informatics from the Technical University of Iasi, Romania, in 2018, respectively 2020. She is currently a Ph.D. student in a joint program between University of Zaragoza, Spain and Technical University of Iasi, Romania. In the academic year 2021-2022 she benefits from an ERASMUS+ program, as well as a 3-month research stay at KTH Royal Institute of Technology in Stockholm, Sweden. Her research interests include path planning and task assignment for multi-agent systems based on discrete event systems and high-level specifications.

Sofia Hustiu is a teaching assistant at Technical University of Iasi, Romania, for the following fields: introduction in control system theory, statistics and path planning strategies for multi-agent systems.
\end{IEEEbiography}

\begin{IEEEbiography}[{\includegraphics[width=1in,height=1.25in,clip,keepaspectratio]{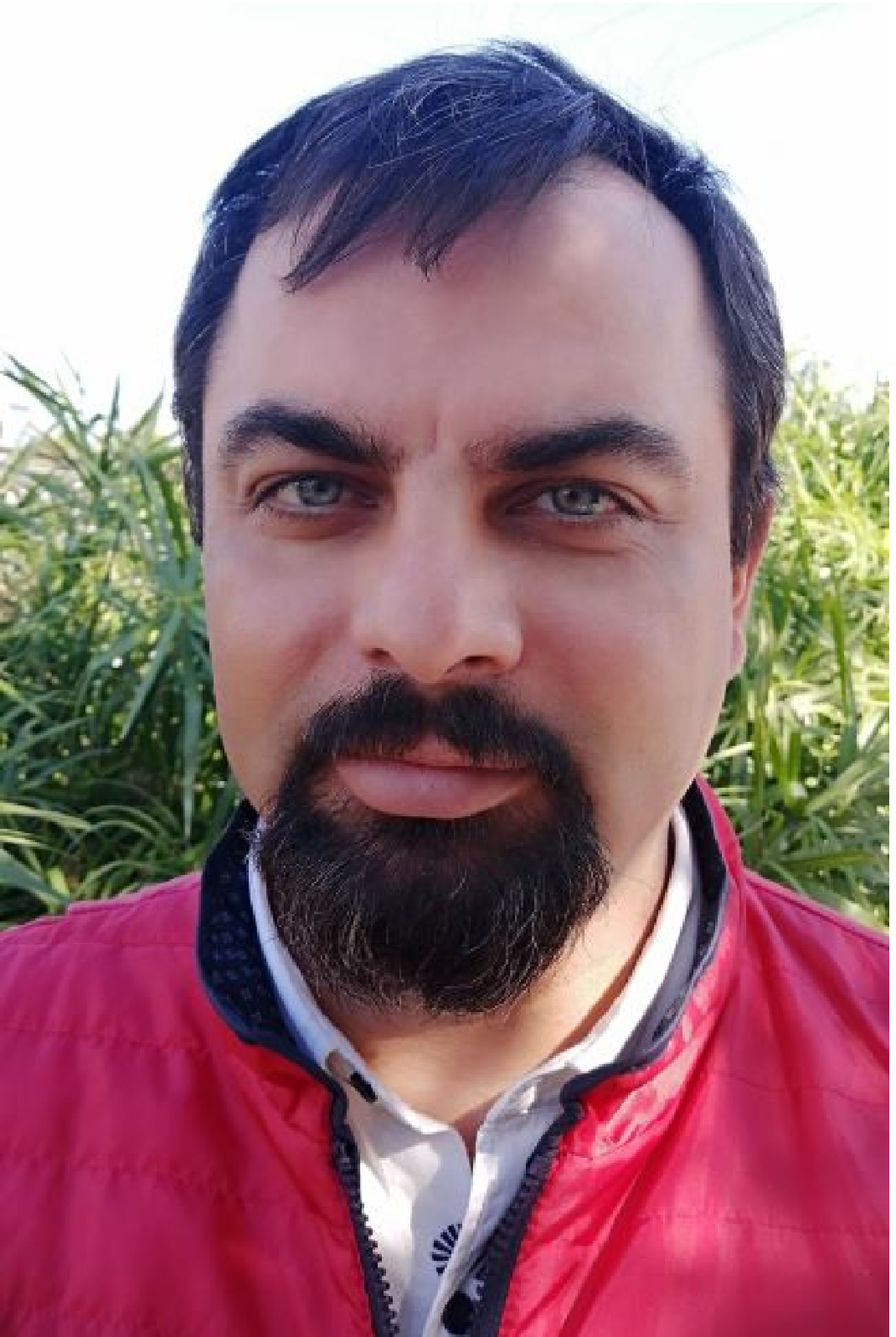}}]{Cristian Mahulea} received the B.S. and M.Sc. degrees in control engineering from the "Gheorghe Asachi" Technical University of Iasi (Romania), in 2001 and 2002, respectively, and Ph.D. degree in systems engineering from the University of Zaragoza (Spain) in 2007. He is an Associate Professor and head of the Department of Computer Science and Systems Engineering at the University of Zaragoza, Spain. His research interests include discrete event systems, hybrid systems, automated manufacturing, Petri nets, mobile robotics and healthcare systems. He  participated in the development of \textbf{Petri Net Toolbox} for the simulation, analysis and synthesis of discrete-event systems modeled with discrete Petri nets and \textbf{RMTool} for path planning and motion control of mobile robots. 

He was visiting professor at the University of Cagliari (Italy) with the Department of Electrical and Electronic Engineering during five months in 2008 and 2010 and a Visiting Researcher at the University of Sheffield (U.K.), University of Cagliari (Italy), Boston University (USA) and ENS Paris-Saclay (France). He has been General Chair of ETFA’2019 being Program Committee chair of ETFA’2017 and ETFA’2018. As editorial activities, Cristian was Associate Editor of IEEE Transactions on Automation Science and Engineering (TASE), and currently he is an Associate Editor of IEEE Control Systems Letters (L-CSS) and of IEEE Transactions on Automatic Control (TAC).
\end{IEEEbiography}

\begin{IEEEbiography}[{\includegraphics[width=1in,height=1.25in,clip,keepaspectratio]{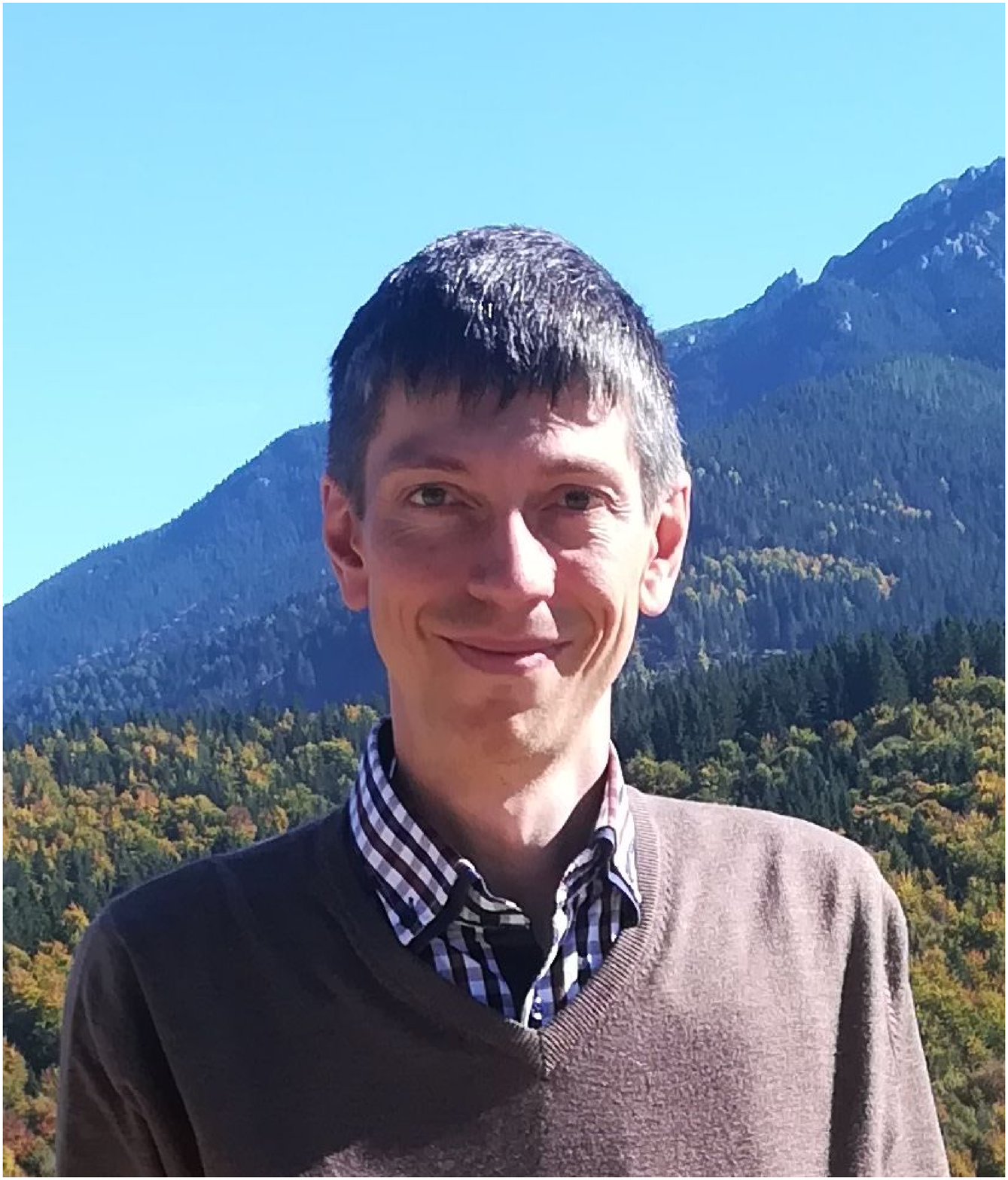}}]{Marius Kloetzer} received the B.S. and M.Sc. degrees in computer science from the Technical University of Iasi, Romania, in 2002 and 2003, respectively, and the Ph.D. degree in systems engineering from Boston University, MA, USA, in 2008. He is currently a Full Professor with the Technical University of Iasi, Romania. His
research interests include formal tools for discrete event systems with applications in motion planning for mobile robots.

Marius Kloetzer was a visiting researcher at Ghent University, Belgium, and at University of Zaragoza, Spain. He has been Organizing Committee chair at ICSTCC'2017 and Work-in-Progress co-chair at ETFA'2019.
\end{IEEEbiography}

\begin{IEEEbiography}[{\includegraphics[width=1in,height=1.25in,clip,keepaspectratio]{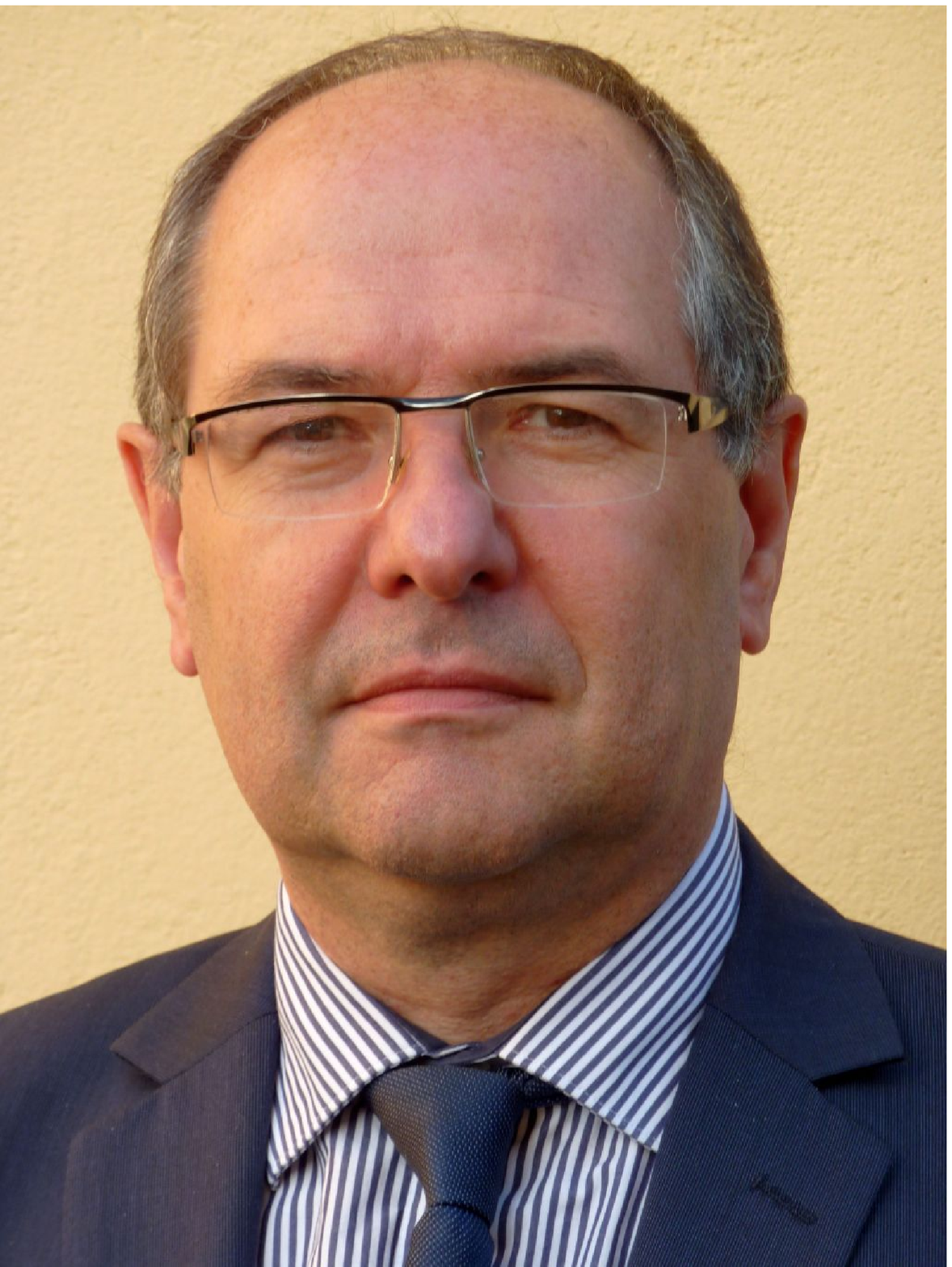}}] {Jean-Jacques Lesage} received the Ph.D. degree from the Ecole Centrale de Paris, France, and the Habilitation à Diriger des Recherches from the University of Nancy, France, in 1989 and 1994 respectively. Currently, he is Professor of Automatic Control with the Ecole Normale Supérieure Paris-Saclay, France.
His research interests include formal methods and models for identification, analysis and diagnosis of Discrete Event Systems, as well as applications to manufacturing systems, network automated systems, energy production, and more recently ambient assisted living.
\end{IEEEbiography}

\end{document}